\documentclass[runningheads]{llncs}

\usepackage{eccv}

\usepackage{eccvabbrv}

\usepackage{graphicx}
\usepackage{booktabs}
\usepackage{multirow}
\usepackage{subcaption}
\usepackage{placeins}

\usepackage[accsupp]{axessibility}  

\usepackage{hyperref}

\usepackage{orcidlink}

\usepackage{tabularray}
\usepackage{xfrac}
\usepackage{siunitx}
\usepackage{mathtools}

\usepackage[normalem]{ulem}
\robustify\bfseries
\robustify\uline

\newcommand{\R}{\mathbb{R}}
\newcommand{\transpose}{\text{t}}
\newcommand{\mat}[1]{\vec{#1}}

\DeclareMathOperator{\first}{\mat{I}}
\DeclareMathOperator{\second}{\mat{I\!I}}
\DeclareMathOperator{\normalize}{normalize}

\usepackage{color}
\usepackage{xcolor}
\definecolor{darkred}{rgb}{0.6,0,0}
\definecolor{darkgreen}{rgb}{0.1,0.3,0.1}
\definecolor{darkorange}{rgb}{0.6,0.4,0.2}
\definecolor{darkpurple}{rgb}{0.4,0.1,0.4}
\definecolor{lightblue}{rgb}{0.5,0.5,0.75}
\definecolor{blue}{rgb}{0,0,0.75}

\def\cpp{{C\nolinebreak[4]\hspace{-.05em}\raisebox{.4ex}{\tiny\bf ++}}}

\newcommand{\vertices}{\mathcal{V}}
\newcommand{\edges}{\mathcal{E}}
\newcommand{\faces}{\mathcal{F}}
\newcommand{\pixels}{\mathcal{P}}
\newcommand{\Star}{\mathcal{S}}

\newcommand{\high}{\ensuremath{0.1\,\text{mm}}}
\newcommand{\medium}{\ensuremath{0.3\,\text{mm}}}
\newcommand{\low}{\ensuremath{1\,\text{mm}}}

\usepackage{scalerel}

\usepackage[draft]{changes}

\begin{document}

\title{An Adaptive Screen-Space Meshing Approach for Normal Integration} 


\author{Moritz Heep\orcidlink{0009-0009-3760-8371} \and
Eduard Zell\orcidlink{0009-0007-3467-9890}} 

\authorrunning{M.~Heep and E.~Zell}

\institute{University of Bonn, Bonn, Germany\\
\href{https://moritzheep.github.io/adaptive-screen-meshing/}{moritzheep.github.io/adaptive-screen-meshing}}

\maketitle

\begin{abstract}
    Reconstructing surfaces from normals is a key component of photometric stereo. This work introduces an adaptive surface triangulation in the image domain and afterwards performs the normal integration on a triangle mesh. Our key insight is that surface curvature can be computed from normals. Based on the curvature, we identify flat areas and aggregate pixels into triangles. The approximation quality is controlled by a single user parameter facilitating a seamless generation of low- to high-resolution meshes. Compared to pixel grids, our triangle meshes adapt locally to surface details and allow for a sparser representation. Our new mesh-based formulation of the normal integration problem is strictly derived from discrete differential geometry and leads to well-conditioned linear systems. Results on real and synthetic data show that 10 to 100 times less vertices are required than pixels.
    Experiments suggest that this sparsity translates into a sublinear runtime in the number of pixels. For 64 MP normal maps, our meshing-first approach generates and integrates meshes in minutes while pixel-based approaches require hours just for the integration.
    \keywords{Surface Reconstruction \and Normal Integration \and Meshing}
\end{abstract}
\begin{figure}
    \centering
    \begin{subfigure}[t]{.19\linewidth}
        \centering
        \includegraphics[width=\linewidth,trim={0 0 0 0},clip]{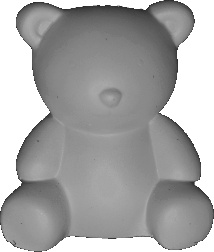}
        \scriptsize{$41512$ Pixels\\Object}
    \end{subfigure}
        \begin{subfigure}[t]{.19\linewidth}
        \centering
        \includegraphics[width=\linewidth,trim={0 0 0 0},clip]{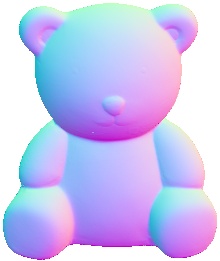}
        \scriptsize{$41512$ Pixels\\Normals}
    \end{subfigure}
    \begin{subfigure}[t]{.19\linewidth}
        \centering
        \includegraphics[width=\linewidth,trim={15 0 15 0},clip]{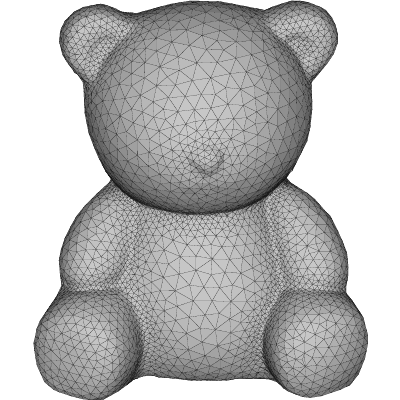}
        \scriptsize{$2523$ Vertices\\@~\high}
    \end{subfigure}
    \begin{subfigure}[t]{.19\linewidth}
        \centering
        \includegraphics[width=\linewidth,trim={15 0 15 0},clip]{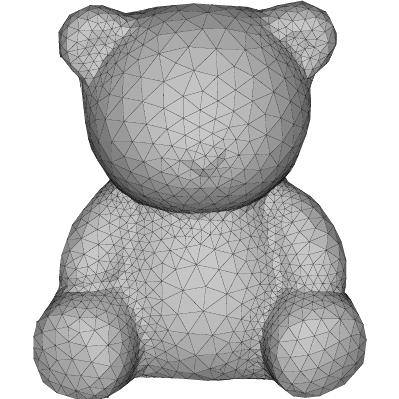}
        \scriptsize{$1118$ Vertices\\@~\medium}
    \end{subfigure}
    \begin{subfigure}[t]{.19\linewidth}
        \centering
        \includegraphics[width=\linewidth,trim={15 0 15 0},clip]{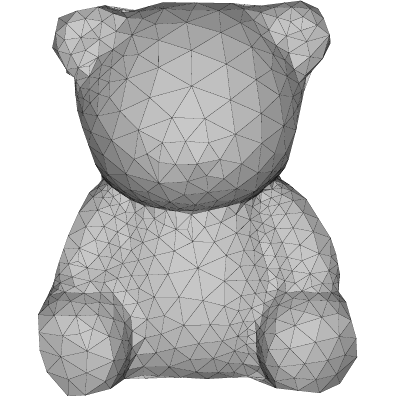}
        \scriptsize{$529$ Vertices\\@~\low}
    \end{subfigure}
\caption{Our screen-space remeshing pipeline decimates smooth, featureless areas efficiently before the normal integration while preserving high-frequency details. Depicted results illustrate high, mid and low-resolution triangulations. Wireframes are rendered as vector graphics for closer examination.}
    \label{fig:not_a_teaser}
\end{figure}
\section{Introduction}
\label{sec:introduction}
Normal integration is renowned for recovering fine and delicate surface details as a subsequent step to photometric stereo  \cite{Queau:2018} or shape from shading. Increasing the resolution of the normal map improves the accuracy of fine structures. At the same time, smooth, featureless regions will reveal little additional information but increase computational costs. Real-world objects often consist of both fine structures and featureless regions and one must choose between sufficient image resolution and reasonable computational cost. In general, to resolve structures half the size, \emph{both} the image height and width must be doubled. This quadratic growth in the number of variables has a significant impact on the runtime of the normal integration and makes regular pixel grids an increasingly inefficient geometric representation.
\par
In practice, many 3D reconstruction pipelines operate at high resolutions and switch to a sparser triangle mesh in the last processing step \cite{Schonberger:2016, Schonberger:2016a, Griwodz:2021}. However, this ignores the underlying problem: all steps up to the final mesh representation are still performed at full resolution, with obvious negative effects on the computational performance.  In contrast to previous work \cite{Gan:2018, Gotardo:2015}, we do not aim to speed up computations in the pixel domain. Instead, we introduce a flexible and locally adaptive triangle mesh \emph{before} the normal integration and solve the problem at its origin, see \cref{fig:not_a_teaser}.
\par
To the best of our knowledge and despite the rich literature on normal integration over the last 30 years, we appear to be the first to propose normal integration on general triangle meshes. Our main insight is that curvature can be extracted from normal maps and that curvature is sufficient to refine triangle meshes locally and in screen space. In addition, we present a novel formulation of the normal integration for triangle meshes. It is strictly derived from discrete differential geometry and is state-of-the-art in that it avoids the checkerboard artefacts and the Gibbs phenomenon described in \cite{Cao:2021}.
Finally, our meshing-first approach can be easily integrated into most existing photometric stereo pipelines and is well-suited for high-quality reconstructions where we achieve significantly sparser representations.
\begin{figure}[tb]
    \centering
        \begin{subfigure}{.155\linewidth}
            \centering
            \caption*{Pixel-Based}
        \end{subfigure}
        \begin{subfigure}{.155\linewidth}
            \centering
            \caption*{Ours}
        \end{subfigure}
        \begin{subfigure}{.155\linewidth}
            \centering
            \caption*{Pixel-Based}
        \end{subfigure}
        \begin{subfigure}{.155\linewidth}
            \centering
            \caption*{Ours}
        \end{subfigure}
        \begin{subfigure}{.155\linewidth}
            \centering
            \caption*{Pixel-Based}
        \end{subfigure}
        \begin{subfigure}{.155\linewidth}
            \centering
            \caption*{Ours}
        \end{subfigure}
        \\
        \begin{subfigure}{.155\linewidth}
            \centering
            \includegraphics[width=\linewidth,trim={140 90 90 220},clip]{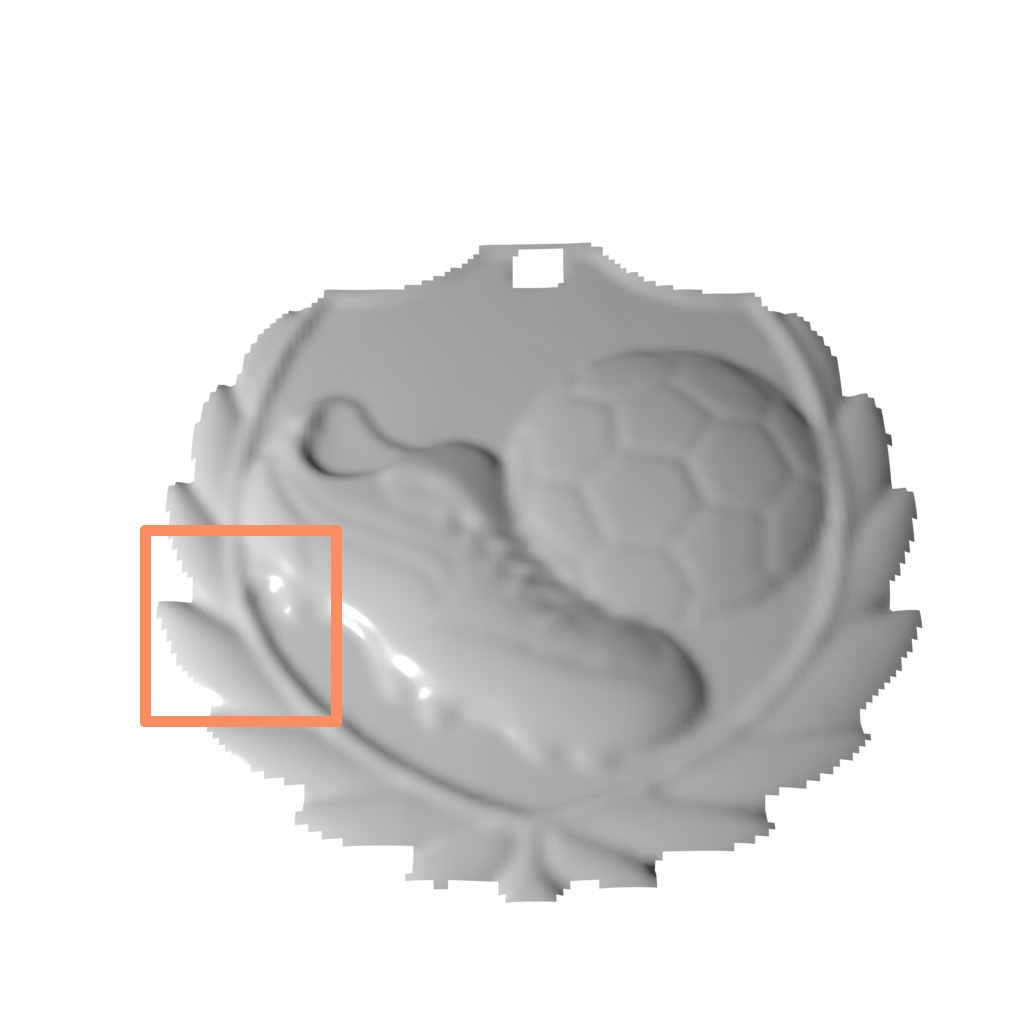} \\
            \includegraphics[width=.9\linewidth]{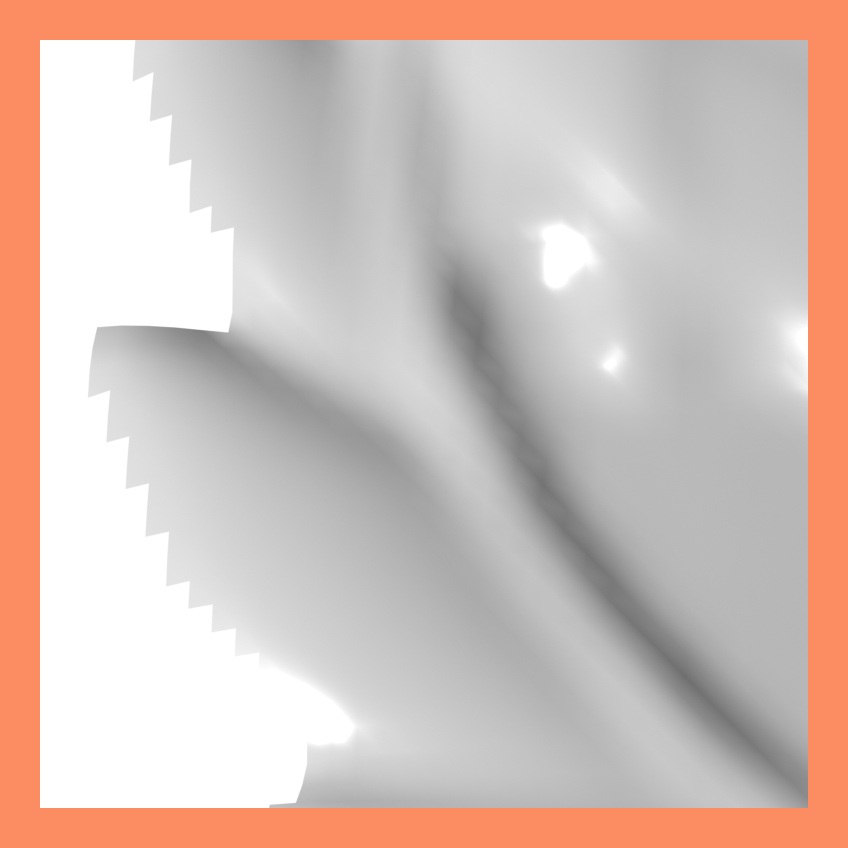}
            \caption*{11k Pixels}
        \end{subfigure}
        \begin{subfigure}{.155\linewidth}
            \centering
            \includegraphics[width=\linewidth,trim={140 90 90 220},clip]{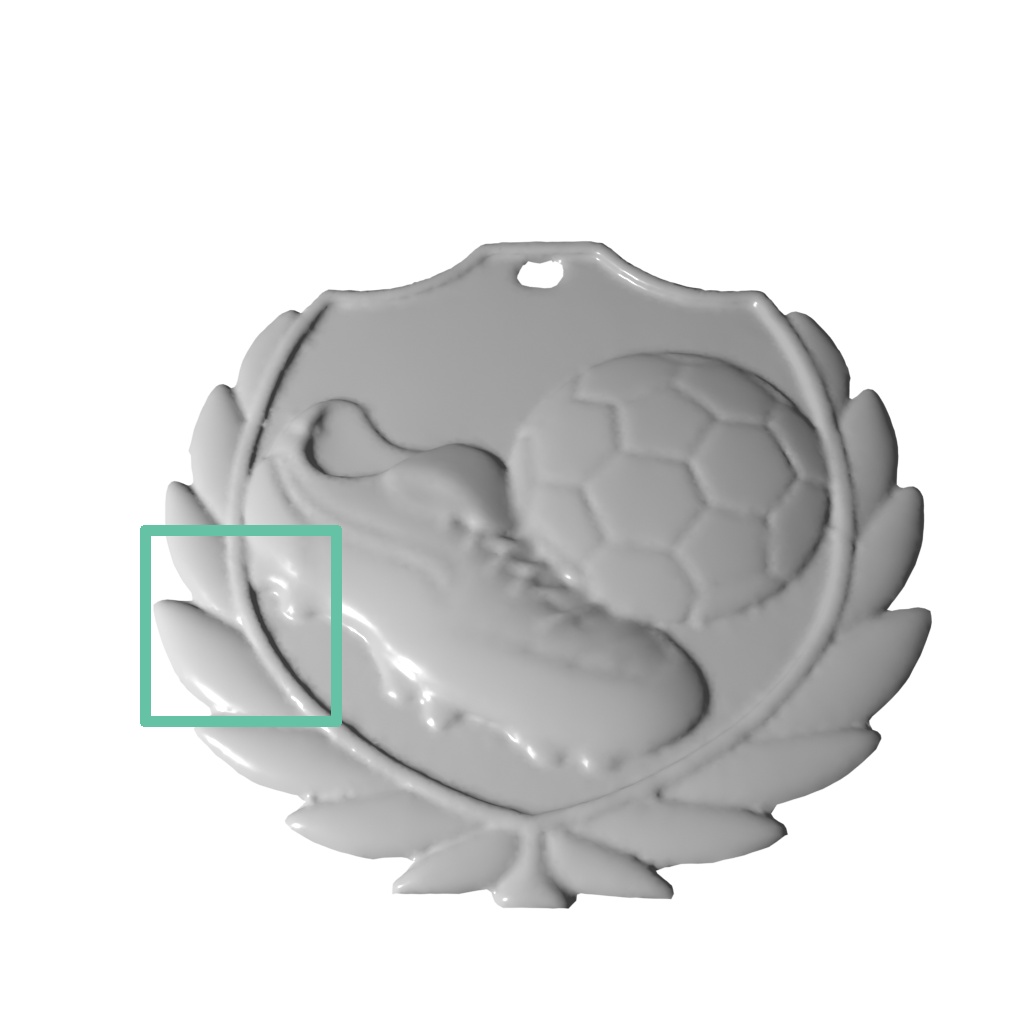} \\
            \includegraphics[width=.9\linewidth]{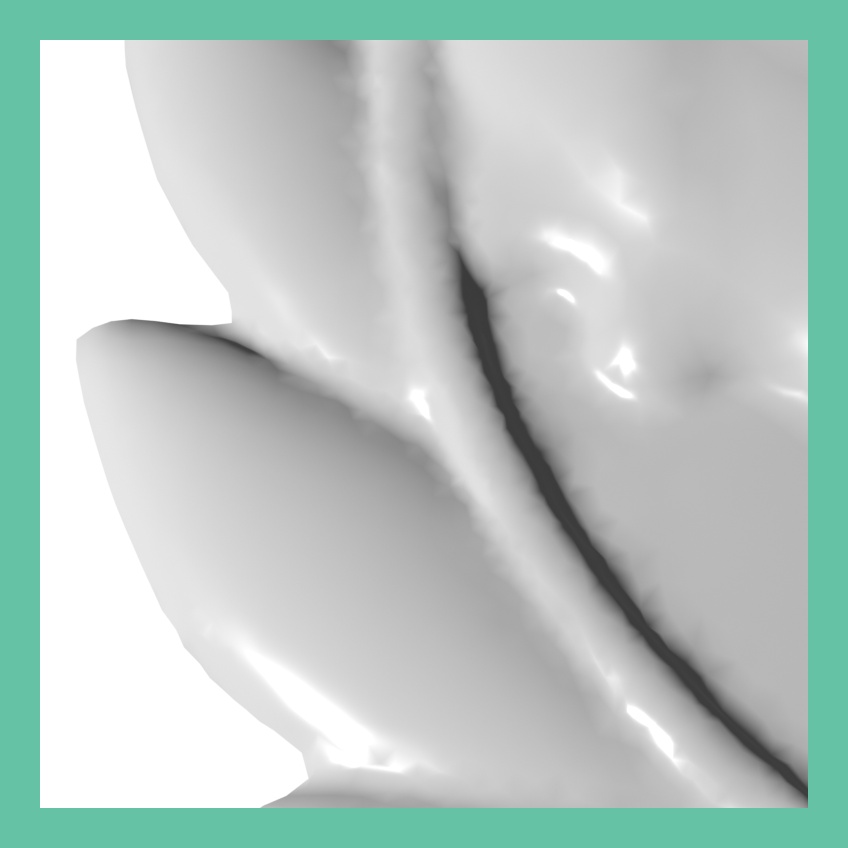}
            \caption*{11k Vertices} 
        \end{subfigure}
        \begin{subfigure}{.155\linewidth}
            \centering
            \includegraphics[width=\linewidth,trim={150 0 150 0},clip]{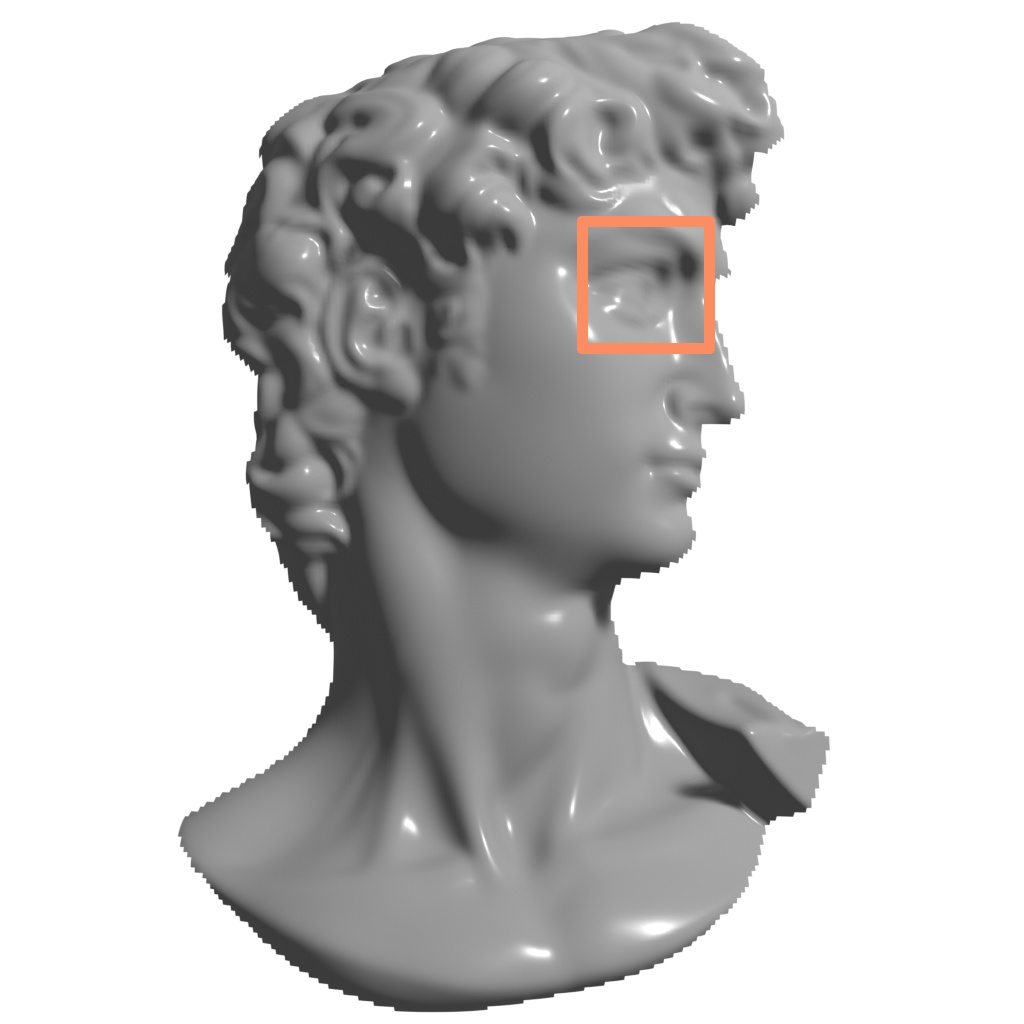} \\
            \includegraphics[width=.9\linewidth]{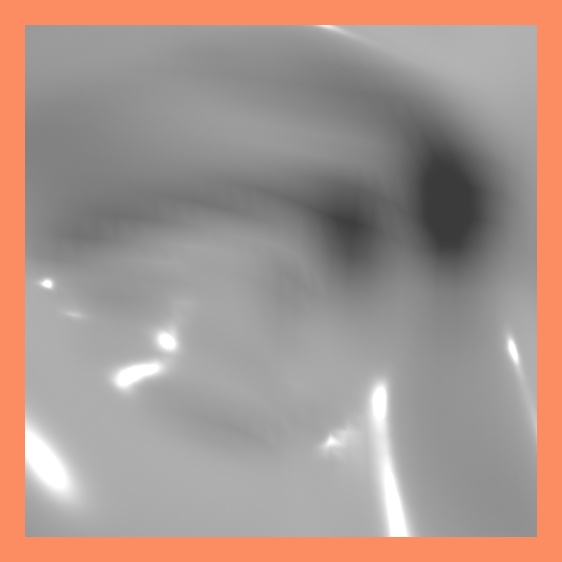}
            \caption*{21k Pixels}
        \end{subfigure}
        \begin{subfigure}{.155\linewidth}
            \centering
            \includegraphics[width=\linewidth,trim={150 0 150 0},clip]{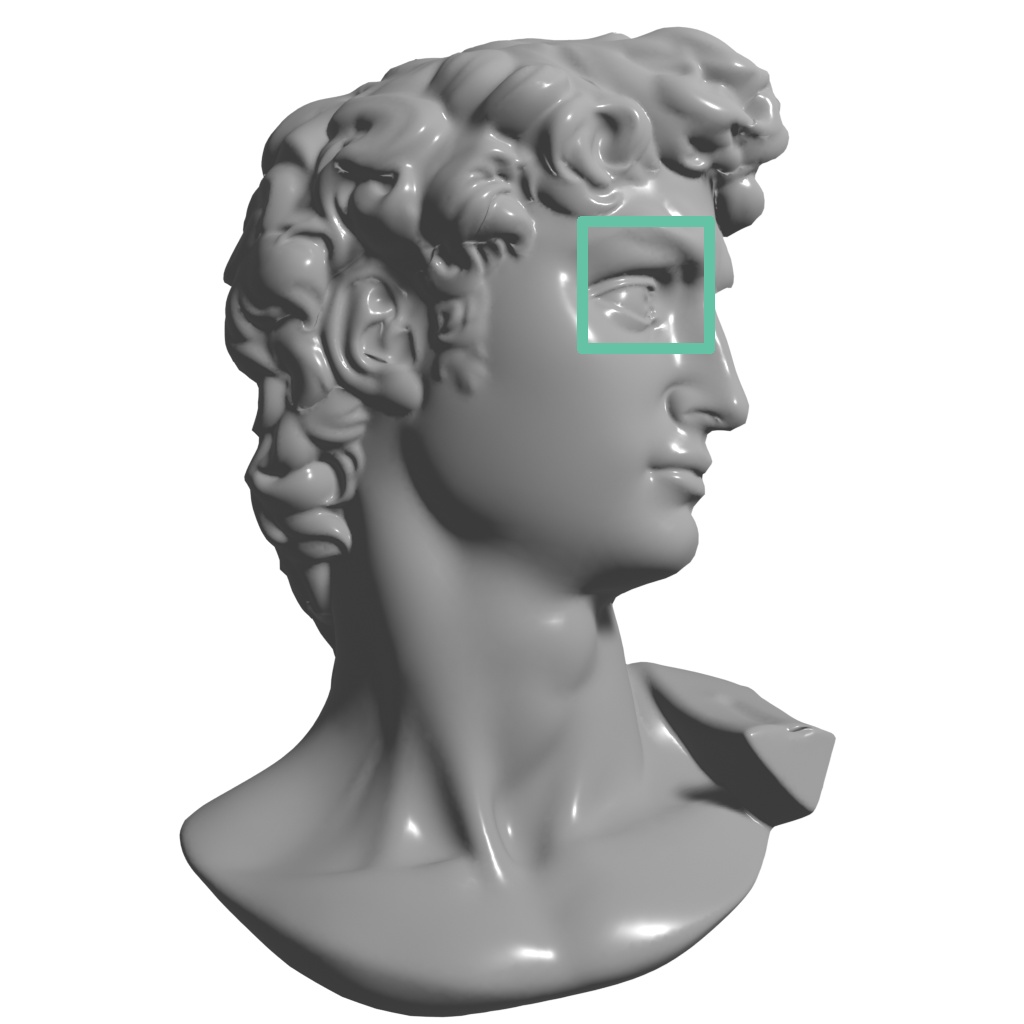} \\
            \includegraphics[width=.9\linewidth]{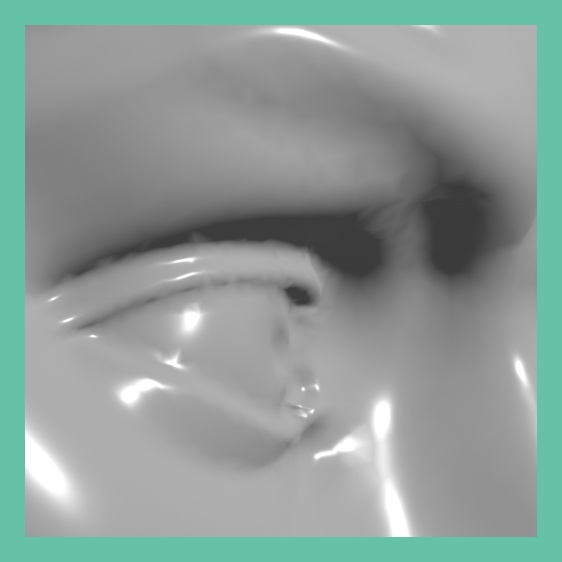}
            \caption*{22k Vertices} 
        \end{subfigure}
        \vrule
        \begin{subfigure}{.155\linewidth}
            \centering
            \includegraphics[width=\linewidth,trim={150 0 150 0},clip]{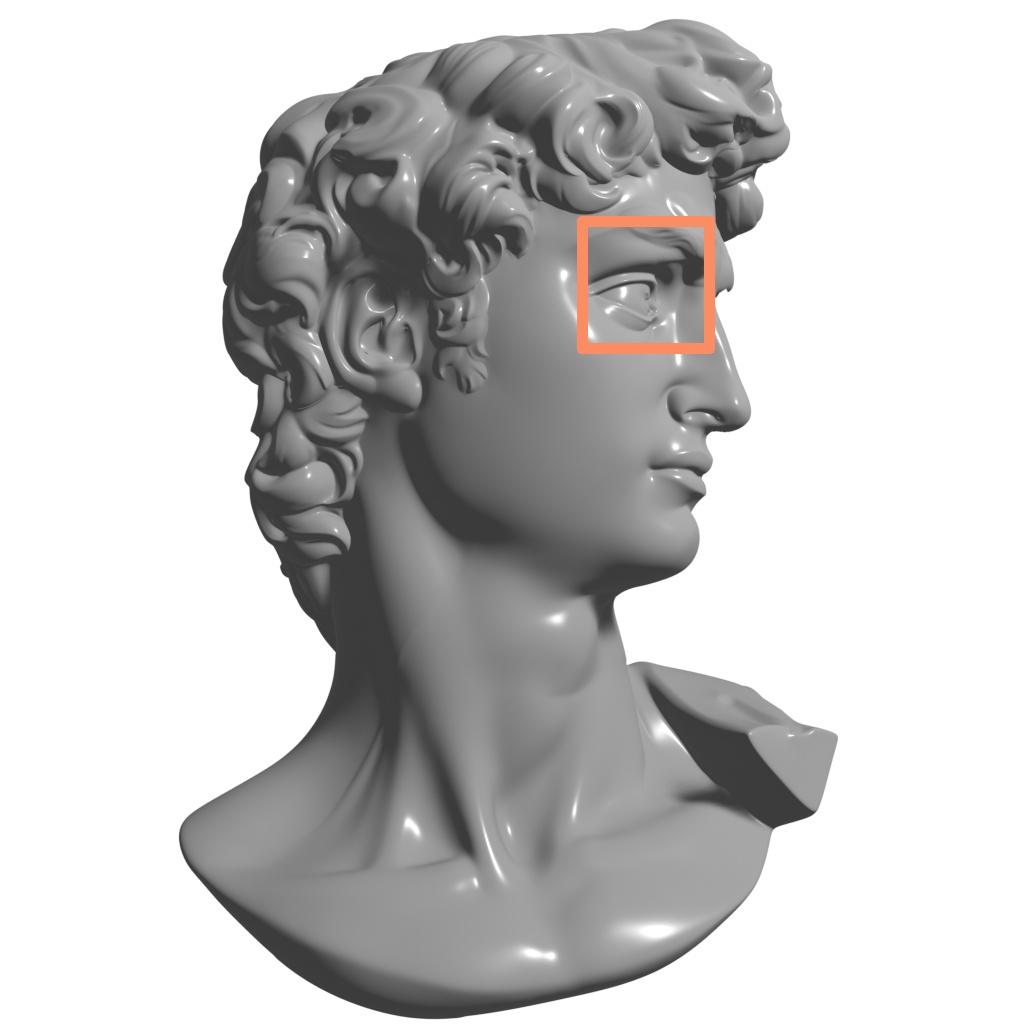} \\
            \includegraphics[width=.9\linewidth]{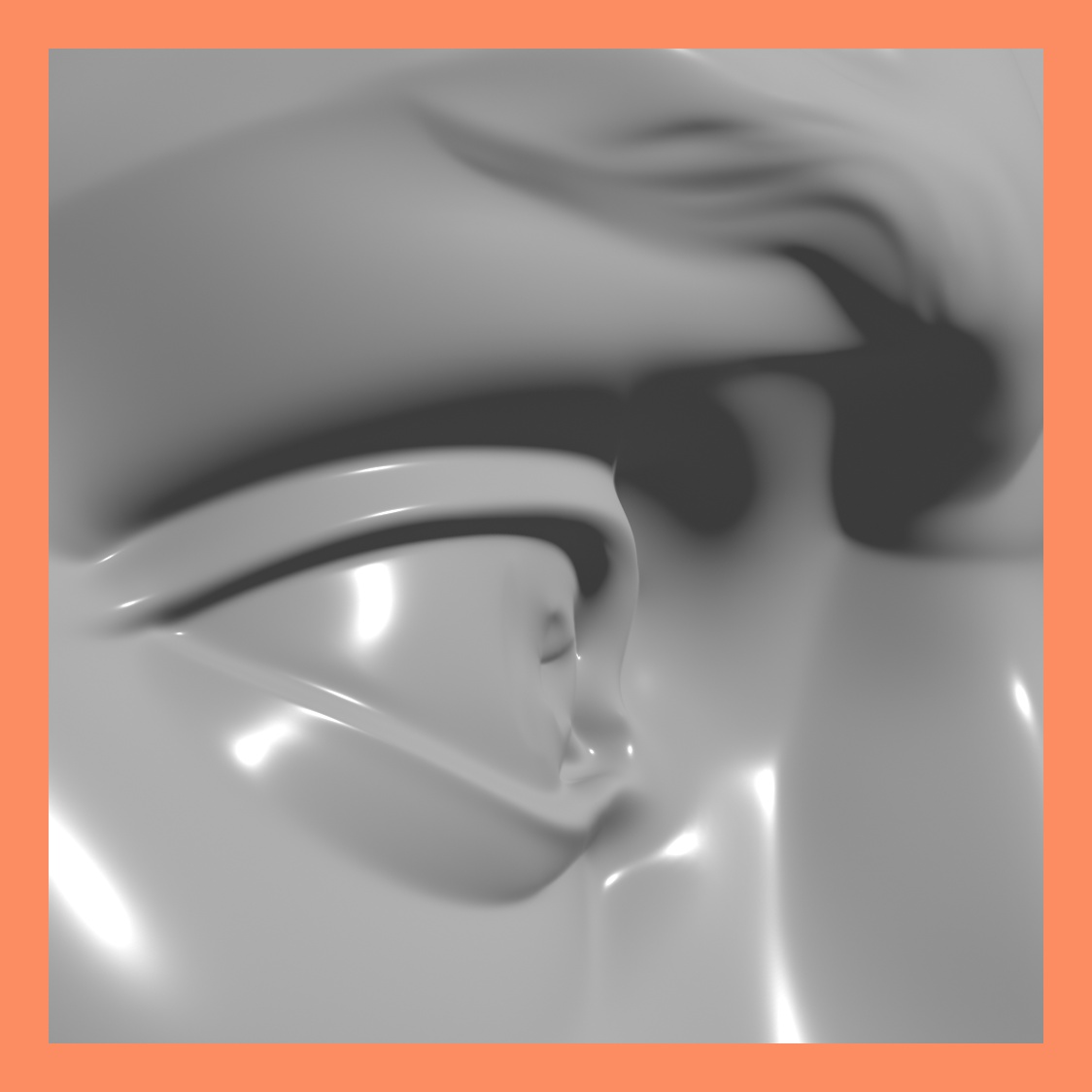}
            \caption*{After 3h}
        \end{subfigure}
        \begin{subfigure}{.155\linewidth}
            \centering
            \includegraphics[width=\linewidth,trim={150 0 150 0},clip]{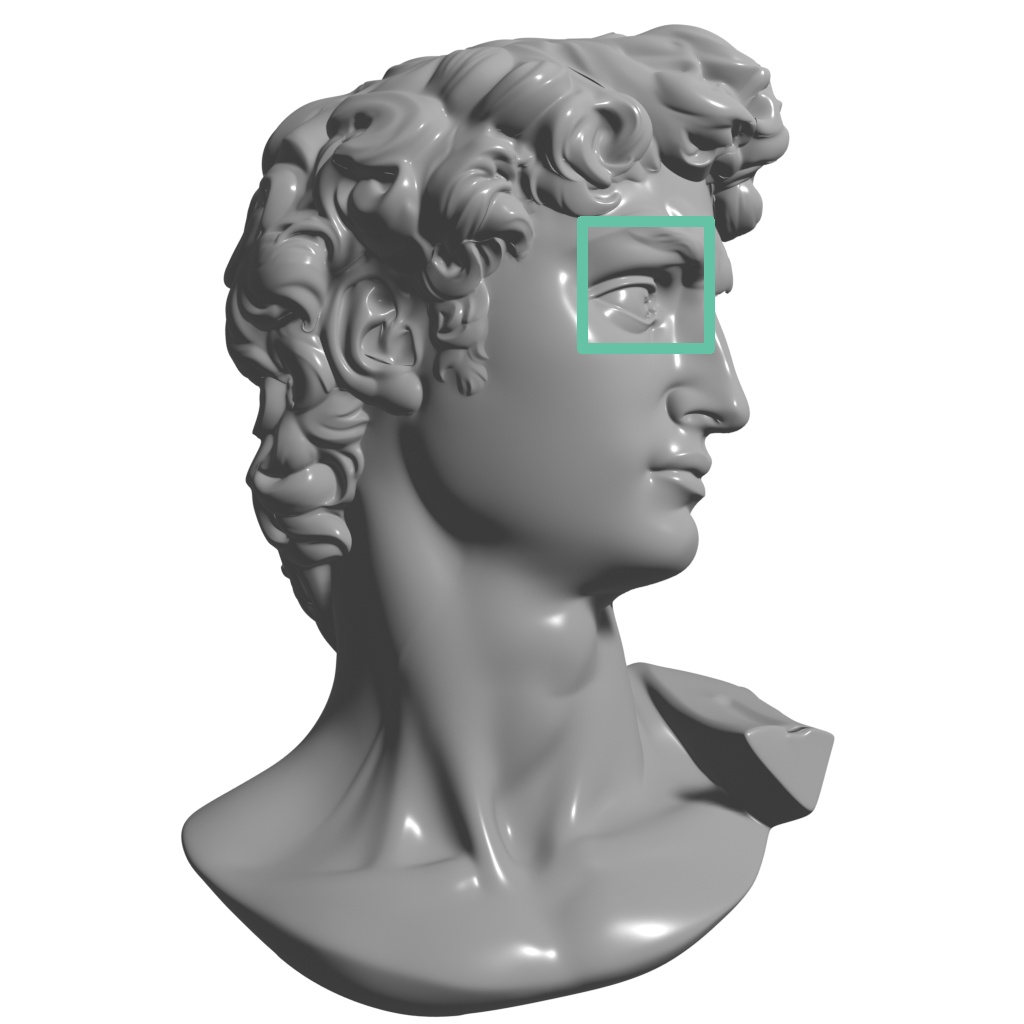} \\
            \includegraphics[width=.9\linewidth]{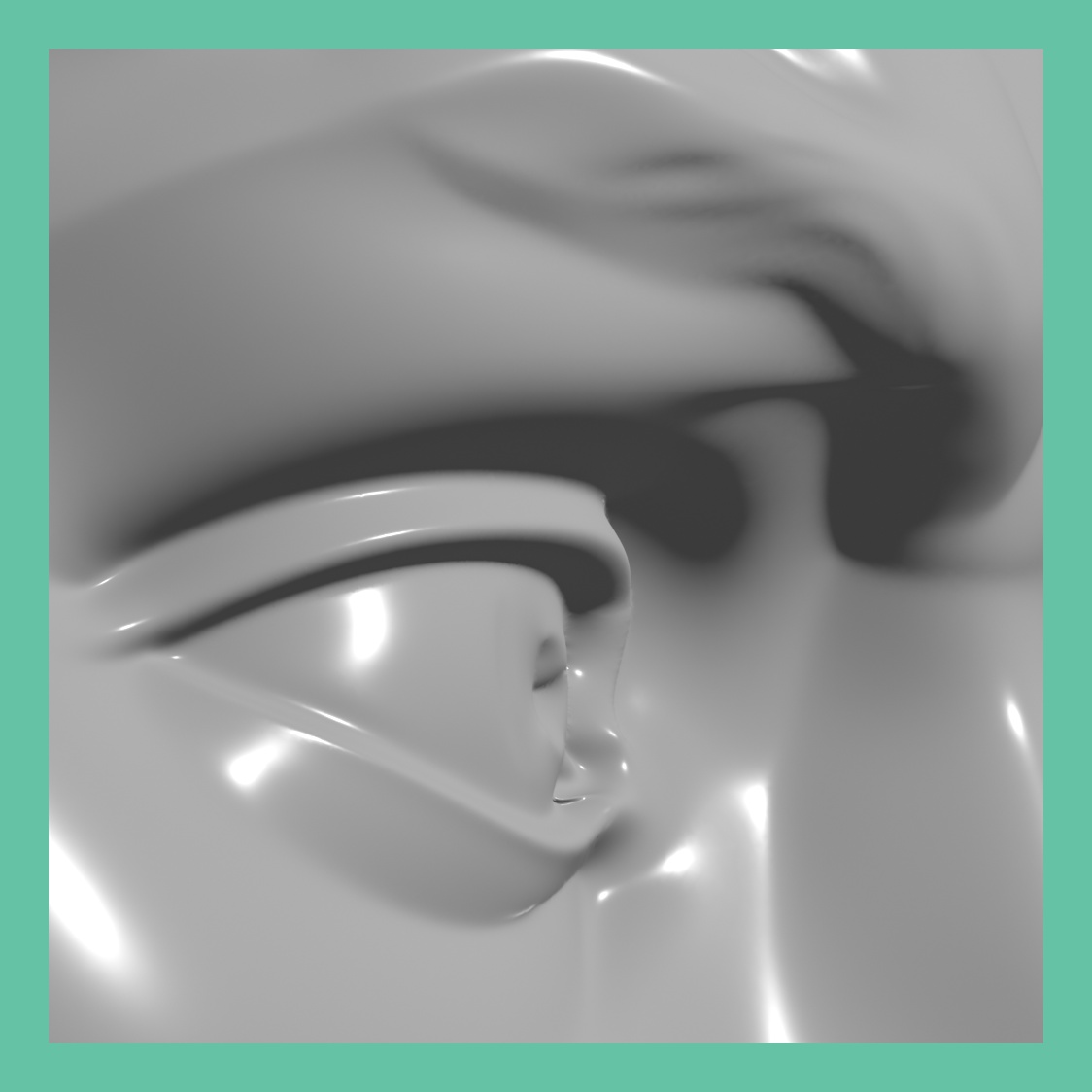}
            \caption*{After 5min}
        \end{subfigure}
    \caption{Left: Using the same number of variables, our meshing-first method achieves higher geometric fidelity. Right: Integration of an $8192^2$ normal map. The pixel-based approach requires around 3 hours while our meshing-first method generates comparable results in only 5 minutes.}
\end{figure}
\section{Related Work}
\label{sec:related_work}
In this section we will focus on dedicated work on normal integration and (re)meshing. We refer the reader to \cite{Ackermann:2015} for a starting point on photometric stereo and how to obtain the normal maps, required by our method.
\par
\begin{figure}[b]
    \centering
    \begin{subfigure}{.49\linewidth}
        \includegraphics[width=\linewidth,trim={150 120 200 270},clip]{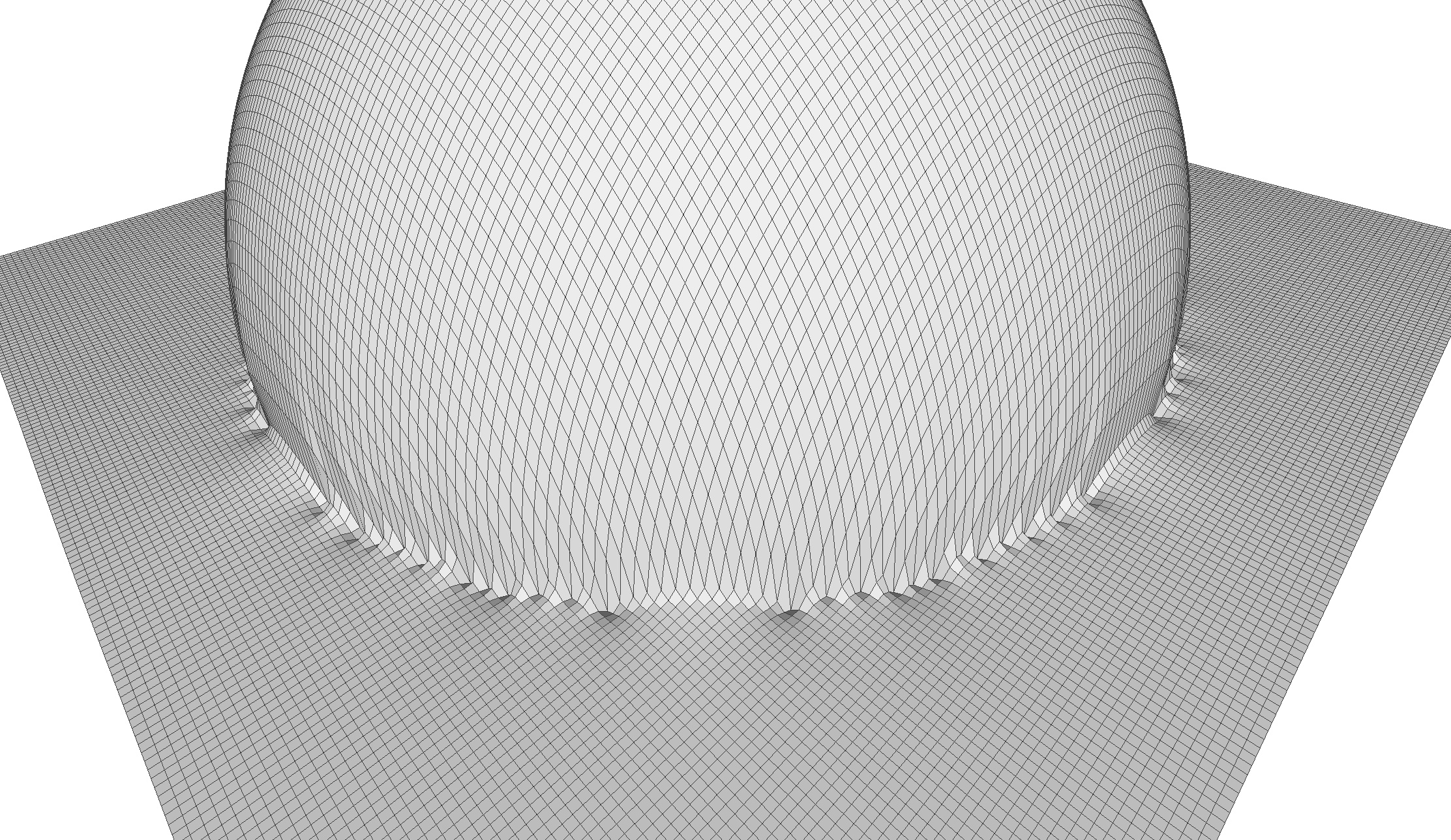}
        \caption{Gradient Formulation \cite{Durou:2007}}
    \end{subfigure}
    \begin{subfigure}{.49\linewidth}
        \includegraphics[width=\linewidth,trim={150 120 200 270},clip]{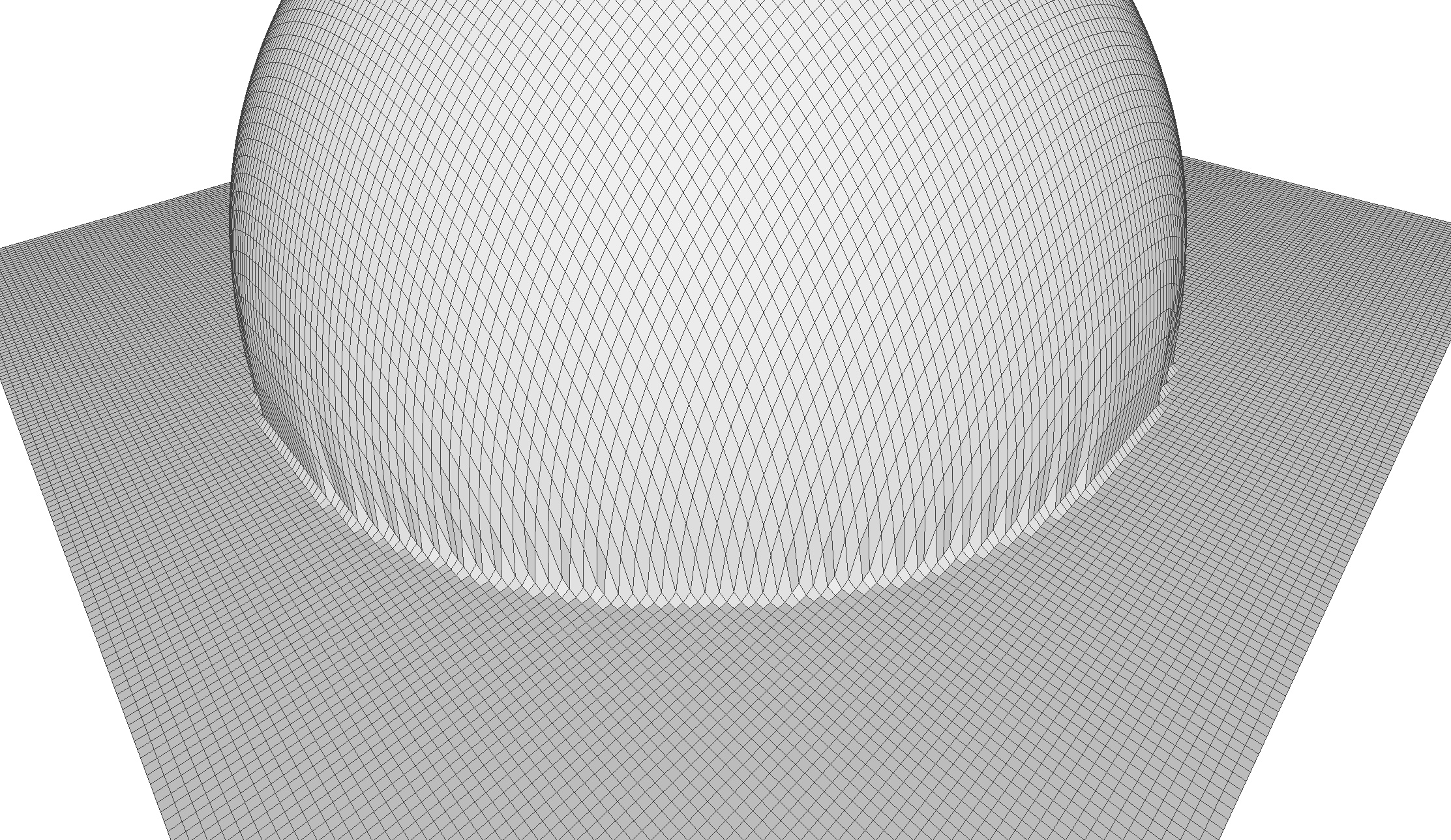}
        \caption{Normal Formulation \cite{Heep:2022}}
    \end{subfigure}
    \caption{Normal integration results with (\emph{left}) and without (\emph{right}) the Gibbs phenomenon.}
    \label{fig:gibbs_phenomenon}
\end{figure}

Most state-of-the-art normal integration methods are so-called \emph{variational methods}, which find a depth map by minimizing an $L^2$ functional containing the difference between the actual depth-map gradients and the observed gradients, \eg from photometric stereo. Broadly, there exist two types of variational methods: Either, functional analysis is used to derive a Poisson equation which is then discretized and solved \cite{Horn:1986} or the functional itself is discretized \cite{Durou:2007}. Both cases lead to a linear system of equations. Methods using the more robust $L^1$ norm have been proposed but are computationally more involved \cite{Du:2007}. More recently, authors have raised concerns about checkerboard artefacts \cite{Zhu:2020} and the Gibbs phenomenon \cite{Cao:2021} (\cref{fig:gibbs_phenomenon}) occurring in the discretized functional setting. Still, these problems are not innate to the variational approach: \cite{Cao:2022} and \cite{Heep:2022} independently proposed a functional using normals over gradients in the pixel-based integration setting. This functional avoids the artefacts mentioned above without resorting to larger stencils for the partial derivatives \cite{Zhou:2020} or introducing additional variables \cite{Cao:2021}. Our method starts with the same functional but discretizes it for general triangle meshes. An overview of variational methods can be found in \cite{Queau:2018}.
\par
Regarding non-variational approaches, Xie \etal \cite{Xie:2014, Xie:2015} consider the regular pixel grid as a quad mesh. They alternate between tilting these quads to align them with the normal directions and glueing adjacent quads into a continuous or even discontinuous \cite{Xie:2019} surface. Although superficially mesh-based, this approach ultimately operates on a pixel-by-pixel basis. Despite being non-variational, Xie \etal's approach still suffers from the Gibbs phenomenon around sharp corners. Cao \etal \cite{Cao:2021} fix the Gibbs phenomenon by optimizing pixel-wise depth values to form a set of interlocking planes defined by the normals. 
\par
In the depth-from-stereo context, image pyramids \cite{Gotardo:2015} or iteratively refined grids of deformation points \cite{Gan:2018} have been proposed to exploit locally similar depth values in smooth regions to speed up reconstruction. While these approaches progressively refine their grid, the grid remains regular and refinement is applied equally over the entire image. In contrast, we adaptively coarsen our grid to allocate computational resources where they are most needed.
\par
The triangulation and remeshing of 2D and 3D surfaces is an intensively studied area \cite{Khan:2020} with varying objectives targeting different applications. For rendering and computer-aided design, the overall shape should be preserved and consist of a minimal amount of vertices \cite{Garland:1997, Yi:2018, Zhao:2023}. However, for numerical simulations like mechanical stress, heat transfer, etc., these very sparse meshes are usually inadequate and more regular triangulations are preferred. Here, Centroidal Voronoi Tesselations (CVT) \cite{Du:1999} as well as Optimal Delaunay Triangulations (ODT) \cite{Chen:2004} are widely used. Both approaches generate uniform or isotropic meshes where each vertex is the (weighted) centroid of some local neighbourhood. The isotropic variants are typically controlled by an application-dependent density. 
Densities based on maximum absolute curvature are popular to balance geometric faithfulness and regularity of the mesh \cite{Alliez:2005, Chen:2012}. An argument for this choice can be found in Dunyach \etal \cite{Dunyach:2013} where a relation between curvature and approximation error is derived. All these approaches have in common that they operate on an existing 3D geometry. In contrast, our approach generates a geometry-adapted triangle mesh from the normal maps alone. We extend the approach of Dunyach \etal \cite{Dunyach:2013} because it directly controls the approximation error. Furthermore, the resulting isotropic meshes increase the numerical stability of our mesh-based normal integration.
\par
Besides surface reconstruction, applications of triangle meshes in 2D include image vectorization \cite{Liao:2012}, although conversion to parametric curves is often preferred in this context \cite{Ma:2022, Dziuba:2023}. The key difference is that image vectorization is based on colours, while our approach approximates the triangulation of a 3D object in 2D screen space.
\section{Definitions, Normals and Curvature} 
\label{sec:preliminaries}
In this section, we will briefly summarize some fundamental concepts in differential geometry and the relation between surfaces, normals and curvatures. This knowledge will be required to derive our solutions for screen-space meshing in \cref{sec:screen_remeshing} and mesh-based integration in \cref{sec:mesh_integration}. 
\par
For a single view, the 3D reconstruction problem can be formulated as finding a projection
\begin{align}
    \phi:\Omega\rightarrow\R^3
\end{align}
from the image foreground $\Omega\subset\R^2$ into 3D space. We write $\vec{u}=(u,v)^\transpose$ for screen coordinates and $\vec{x}=(x,y,z)^\transpose$ for object coordinates. The projection is usually parametrized by a depth map $h:\Omega\rightarrow\R$ as
\begin{align}
    \vec{x}:=\phi(\vec{u})=\big(u,v,h(\vec{u})\big)^\transpose
\end{align}
in the orthographic case and
\begin{align}
    \vec{x}:=\phi(\vec{u})=\vec{r}(\vec{u})\cdot h(\vec{u})
\end{align}
in the projective case where $\vec{r}:=\mat{C}^{-1}\cdot(u,v,1)$ is the ray given by the camera matrix $\mat{C}$. As $h=z$ for both projections, we will use $z$ synonymously.
\subsection{Normal Integration}
Given a normal map $\vec{n}:\Omega\rightarrow\mathcal{S}^2$, the normal integration problem is finding the depth map $z$ that minimizes
\begin{align}
    E_\text{Int}=\int_\Omega\langle\vec{n},\partial_u\vec{x}\rangle^2+\langle\vec{n},\partial_v\vec{x}\rangle^2\,du\,dv\,.
    \label{eqn:integration_continuous}
\end{align}
This modified Dirichlet energy has only recently been proposed in the context of discontinuity preserving normal integration \cite{Cao:2022} and depth-from-stereo \cite{Heep:2022}. Using normals instead of depth map gradients overcomes the problem of distortions near sharp corners. This phenomenon, also known as the Gibbs phenomenon \cite{Cao:2021}, \cref{fig:gibbs_phenomenon}, is present in traditional variational approaches and is discussed in more depth within the supplementary material. Relying on discrete differential geometry, we derive our novel mesh-based discretization of the modified Dirichlet energy in \cref{sec:mesh_integration}.
\begin{figure}[tb]
    \centering
    \begin{subfigure}{.30\linewidth}
        \includegraphics[width=\linewidth,trim={40 55 465 105},clip]{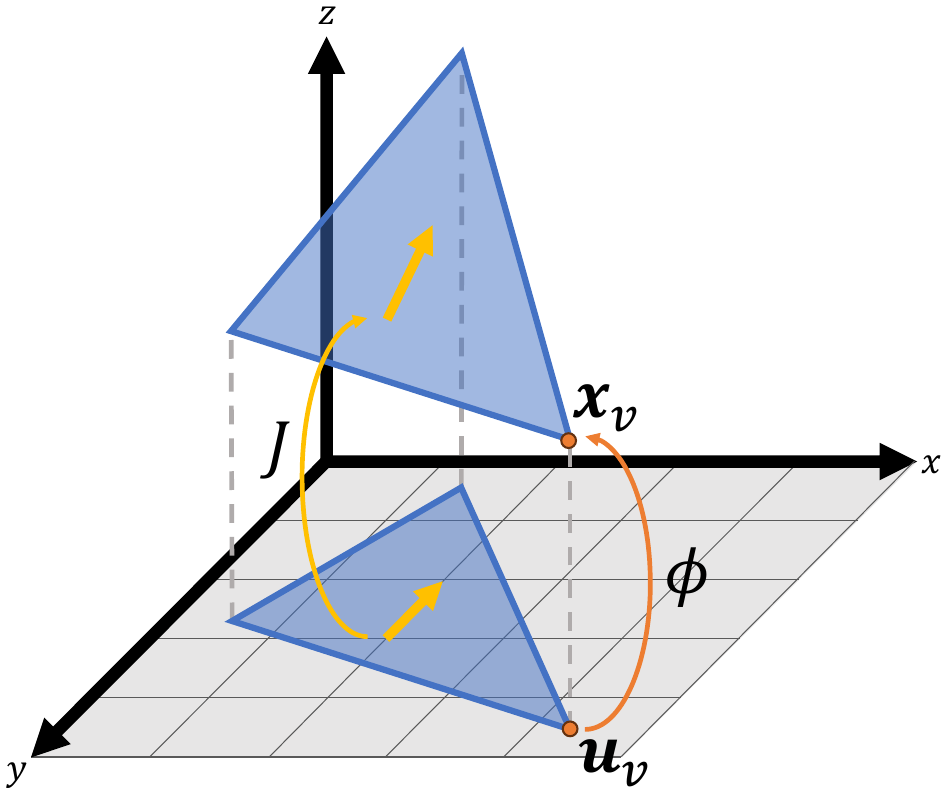}
        \caption{Foreshortening}
        \label{fig:maps}
    \end{subfigure}
    \hspace{20mm}
    \begin{subfigure}{.32\linewidth}
        \includegraphics[width=\linewidth,trim={10 225 390 0},clip]{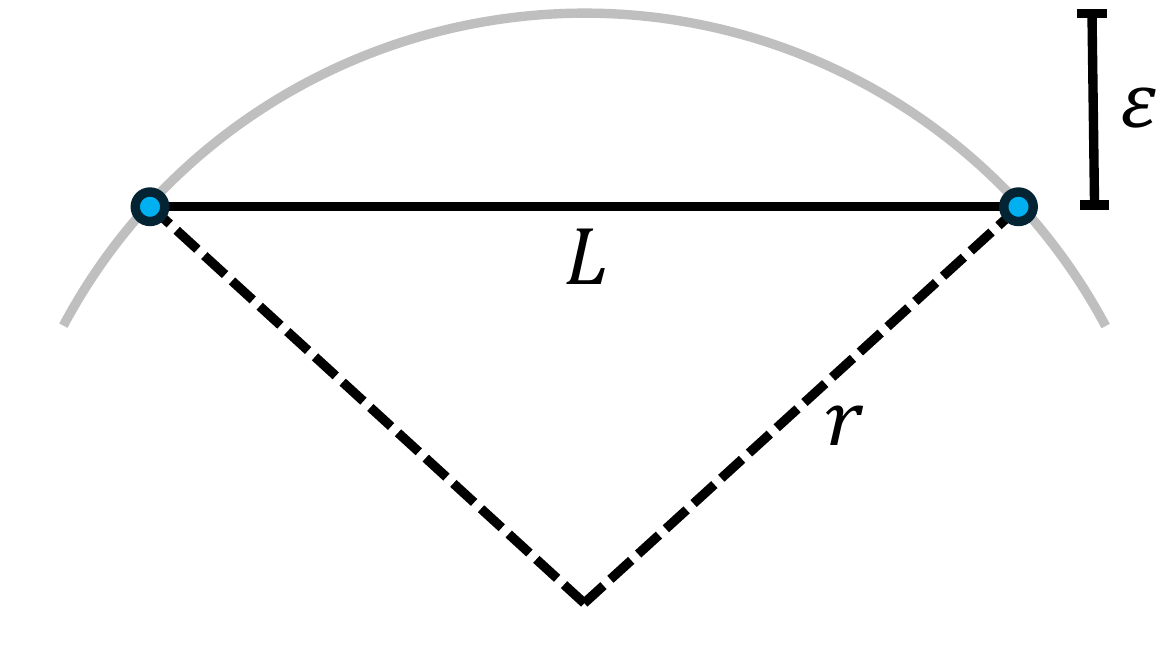}
        \caption{Error Estimate}
        \label{fig:error_estimate}
    \end{subfigure}
    \caption{Left: The projection $\phi$ maps points on the screen into three-dimensional space. Distances on the screen appear shorter than their three-dimensional counterparts (foreshortening). This is captured in the Jacobian $J$ of $\phi$. Right: Illustration on the relationship between the edge lengths and the user-defined approximation error $\epsilon$ as derived in \cite{Dunyach:2013}.}
    \label{fig:foreshortening_epsilon}
\end{figure}
\subsection{Curvature}
\label{sec:curvature}
Strong changes in the normals indicate geometric details. Mathematically speaking, this is described by curvature $\kappa$. Just as normals provide a planar approximation of the surface, so does curvature provide a spherical approximation. The radius of this approximating sphere is exactly $|\kappa|^{-1}$. Curvatures are calculated from the first and second fundamental forms:
\begin{align}
    \label{eqn:fundamental_forms}
    \first_{ij}  &= \partial_i\vec{x}\cdot\partial_j\vec{x}\,, &
    \second_{ij} &= -\partial_i\vec{x}\cdot\partial_j\vec{n}\,.
\end{align}
where $i,j\in\{u,v\}$. The first fundamental form quantifies foreshortening and measures the 3D distance between two points on the 2D screen. The second fundamental form is virtually the gradient of the normal map. However, the second fundamental form itself is subject to foreshortening. Instead, we consider the two eigenvalues $\kappa_i$ and eigenvectors $\vec{v}_i$ of the generalized eigenvalue problem
\begin{align}
    \kappa_i\cdot\first\vec{v}_i=\second\vec{v}_i\,.
    \label{eqn:principal_eigen}
\end{align}
They are called the principal curvatures and principal directions respectively. The addition of the first fundamental form in the generalized eigenvalue problem exactly compensates for foreshortening.
\par
Given a set of surface normals, we can calculate tangents since $\vec{n}\cdot\partial_i\vec{x}=0$. In the orthographic case, this gives
\begin{align}
    \partial_u\vec{x}   &=  \vec{e}_x-\frac{\vec{n}_x}{\vec{n}_z}\cdot\vec{e}_z &
    \partial_v\vec{x}   &=  \vec{e}_y-\frac{\vec{n}_y}{\vec{n}_z}\cdot\vec{e}_z &
    \label{eqn:tangent_orthographic}
\end{align}
where $\vec{e}_x,\vec{e}_y,\vec{e}_z$ are the unit vectors in the respective coordinate axis directions. As no knowledge about the depth map $z$ is required, one can perform all these calculations using normal maps alone. This is the foundation for our screen-space triangulation. 
\par
In the projective case, we get
\begin{align}
    \partial_u\vec{x}   &=  \left(\partial_u\vec{r}-\frac{\vec{n}\cdot\partial_u\vec{r}}{\vec{n}\cdot\vec{r}}\cdot\vec{r}\right)\cdot z   &
    \partial_v\vec{x}   &=  \left(\partial_v\vec{r}-\frac{\vec{n}\cdot\partial_v\vec{r}}{\vec{n}\cdot\vec{r}}\cdot\vec{r}\right)\cdot z
    \label{eqn:tangent_perspective}
\end{align}
where the ray $\vec{r}$ is again given by the camera matrix. Unlike the orthographic case, the tangents depend on the depth map $z$. However, if distances within the object are small compared to the camera-to-object distance, we can adopt the weak perspective setting and set $z$ to the average camera-to-object distance. In the supplementary material, we go into detail on how to handle objects very close to the camera.
\begin{figure}[tb]
    \centering
    \includegraphics[width=\linewidth, trim={0 385 0 70}, clip]{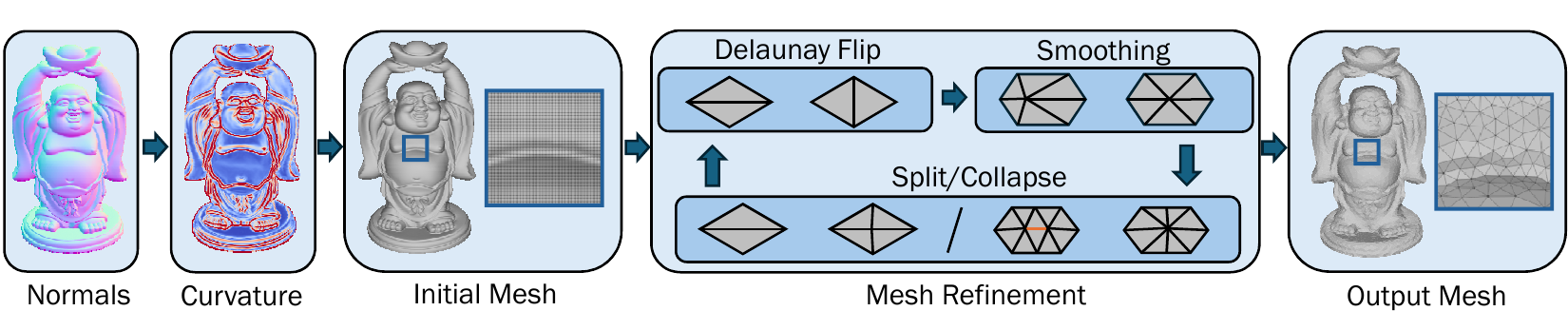}
    \caption{The meshing pipeline: We start by calculating curvature from the normal maps. This curvature is used to derive an optimal edge length. Then, an initial mesh is iteratively refined to move its edge lengths closer to the calculated optimum. The result is a sparse, isotropic mesh.}
    \label{fig:overview_remeshing}
\end{figure}
\section{Screen-Space Meshing}
\label{sec:screen_remeshing}
To create a feature-adaptive mesh before the actual surface integration, we must reliably predict detailed surface areas and adjust the density of the triangle mesh accordingly. 
While mesh simplification algorithms like \cite{Garland:1997} achieve an impressive reduction in vertex number, their sole focus on geometric faithfulness often leads to skinny triangles. As we will see in \ref{sec:mesh_integration}, small angles decrease the numerical stability of the mesh-based integration. In contrast, remeshing algorithms also take mesh quality into account and are much more suited for our use-case.
That being said, existing remeshing methods operate on 3D surfaces and are not directly applicable to our case. Searching for a fast and robust method with only a few parameters, we identified the work by Dunyach \etal \cite{Dunyach:2013} as the most suitable candidate. In their algorithm, all edge lengths are optimized towards an optimal edge length 
\begin{align}
    L=\sqrt{\frac{6\epsilon}{|\kappa|}-\epsilon^2}
    \label{eqn:optimal_length}
\end{align}
that is based on curvature and a user-defined approximation error $\epsilon$, see \cref{fig:foreshortening_epsilon}b. However, there are two obstacles: The first issue is the edge length; lengths on screen are subject to foreshortening, while the optimal edge length is not. They cannot be simply compared without compensation (\cref{sec:measuring_lengths}). The second issue is that we require a vertex-based curvature measure to obtain the optimal lengths. In \cref{sec:curvature}, we discussed how to compute pixel-wise curvatures from normal maps and we will present a solution for computing vertex-based curvatures in screen space in \cref{sec:calculating_curvatures}. Finally, we will discuss the optimization of vertex positions (\cref{sec:tangential_smoothing}). For all these steps, we assume that the screen is already covered by a triangle mesh. In \cref{sec:iterative_mesh_refinement}, we show how we obtain the initial triangulation of the screen and summarize our screen-space algorithm to obtain the final triangulation. An overview of our approach is given in \cref{fig:overview_remeshing}.
\subsection{Measuring Lengths}
\label{sec:measuring_lengths}
Edges in screen-space are subject to foreshortening, \ie appear shorter on screen than in 3D, see \cref{fig:maps}. This foreshortening is exactly described by the first fundamental form $\first$, \cref{eqn:fundamental_forms}. We have shown in \cref{sec:curvature}, how to derive the first fundamental form from normals. To apply this to the mesh setting, we use the pixel-wise normal map to obtain face normals
\begin{align}
    \vec{n}_f:=\normalize\left(\textstyle\sum_{p\in\pixels_f}\vec{n}_p\right)
\end{align}
where $\pixels_f$ contains all pixels in the image that are covered by triangle $f$. As large triangles occur only in regions with low curvature, where normals vary slowly, the variance of $\vec{n}_f$ is small. Plugging this normal into \cref{eqn:tangent_orthographic,eqn:tangent_perspective} respectively, we obtain fundamental forms for each face. With these fundamental forms, the length of an edge $(i,j)\in\edges$ is
\begin{align}
    l_{ij}=\sqrt{\frac{1}{2}\cdot\Big((\vec{u}_i-\vec{u}_j)^t\cdot\first_{f}\cdot(\vec{u}_i-\vec{u}_j)+(\vec{u}_i-\vec{u}_j)^t\cdot\first_{f'}\cdot(\vec{u}_i-\vec{u}_j)\Big)}
\end{align}
where $f,f'$ are the two faces adjacent to the edge and $\vec{u}_i,\vec{u}_j$ are the \emph{screen} coordinates of the edge's endpoints. For boundary edges, there is only one adjacent face and we omit the factor $\sfrac{1}{2}$.
\subsection{Calculating Curvatures}
\label{sec:calculating_curvatures}
To obtain vertex curvatures, we start by calculating maximum absolute curvature $\kappa^{(\text{max})}_p$ for each pixel by solving the eigenvalue problem in \cref{eqn:principal_eigen}. For this, we require the first and second fundamental forms, \cref{eqn:fundamental_forms}. Depending on the projections, we use \cref{eqn:tangent_orthographic,eqn:tangent_perspective} respectively for the surface gradients and finite difference for gradients of the normal maps. We lift the pixel-wise curvatures to the mesh setting via
\begin{align}
    \kappa_v:=\max\bigg(\big\{\kappa^{(\text{max})}_p\big|p\in\pixels_{\Star_v}\big\}\bigg)\,.
\end{align}
where $\Star_v$ is the star of $v$, i.e.\ the union of all triangles touching $v$. Taking the maximum ensures small triangles at sharp corners. Sticking to the work of Dunyach \etal, we use \cref{eqn:optimal_length} to obtain optimal lengths $L_v$ which are nominally vertex properties. The optimal length for an edge is simply the mean of these vertex properties. 
\subsection{Tangential Smoothing}
\label{sec:tangential_smoothing}
To obtain an optimal Delaunay triangulation \cite{Chen:2004} and avoid triangles with small angles, we move each vertex to the centroid of its star, \ie the union of its adjacent faces. In this case, we have to address foreshortening twice. First, the centroid is affected by the triangle size which is typically smaller after projection. Second, the centroid of the star is a weighted sum over the centroids of the faces in the star. The position of the central vertex relative to these face centroids is foreshortened too. Considering both aspects yields a linear system
\begin{align}
    \left(\sum_{f\in\faces_v}\frac{A_f\cdot\sqrt{\det\first_f}}{L_f^2}\cdot\first_f\right)\cdot\vec{u}_v^{(k+1)}=\left(\sum_{f\in\faces_v}\frac{A_f\cdot\sqrt{\det\first_f}}{L_f^2}\cdot\first_f\vec{c}_f^{(k)}\right)
\end{align}
where the areas $A_f$ and centroids $\vec{c}_f$ are measured as they appear on-screen and foreshortening is compensated by the inclusion of the face's first fundamental forms $\first_f$. $L_f$ is simply the average of optimal lengths for the three vertices in $f$.
\subsection{Iterative Mesh Refinement}
\label{sec:iterative_mesh_refinement}
After deriving all measurements for curvature and edge lengths in screen-space, we formulate our screen-based remeshing. We start with a uniform triangulation of the pixel grid, where each foreground pixel is split into two triangles. Following previous remeshing methods \cite{Botsch:2004, Dunyach:2013}, which are visualized in \cref{fig:overview_remeshing}, the mesh is iteratively refined by performing the following steps:
\begin{enumerate}
    \item Collapse short edges and split long edges.
    \item Flip non-Delaunay edges.
    \item Shift vertices (tangential smoothing).
\end{enumerate}
To determine whether an edge is too long or too short, we use the optimal edge lengths and the heuristic constants derived in \cite{Botsch:2004}.
\section{Mesh-Based Integration}
\label{sec:mesh_integration}
For normal integration, previous authors \cite{Queau:2018, Cao:2022} have noted that a unified treatment of the orthographic and perspective case can be achieved by using the functional
\begin{align}
    E_\text{Int}=\int_\Omega \left(\vec{n}\cdot\vec{r}\,\partial_u z +n_x\right)^2+\left(\vec{n}\cdot\vec{r}\,\partial_v z +n_y\right)^2\,du\,dv
    \label{eqn:unified_integration}
\end{align}
where $\vec{r}=\vec{e}_z$ and $z$ the depth map in the orthographic case. In the perspective case, $z$ is the logarithmic depth and $\vec{r}$ is the camera ray given by the intrinsics. Using this unified formulation, we do not differentiate between the orthographic and perspective cases for the rest of this section. The full derivation of the mesh-based normal integration contains a lot of details and the interested reader is referred to the supplementary material. In this section, we will summarize only the main results.
\par
In the mesh-based normal integration, we have one unknown depth value per vertex. Depth values within the triangle are obtained through linear interpolation using barycentric coordinates. By linearity, the derivatives in \cref{eqn:unified_integration} are constant within each face \cite{Botsch:2010} and so are the normals. The integral can then be split into a sum of integrals over faces which can be carried out explicitly. The result is a discretized energy that is quadratic in the vertex depths. Taking the derivative, the optimality condition is
\begin{align}
    \sum_{i\in\vertices_j}\sum_{f\in\faces_{ij}}\cot(\alpha_{f,ij})\cdot\left(2m_f\cdot(z_j-z_i)+b_f\cdot 
    \begin{pmatrix}
        n_x \\  n_y    
    \end{pmatrix}\cdot(\vec{u}_j-\vec{u}_i)\right)=0
    \label{eqn:optimality_condition}
\end{align}
for each vertex $i$. $\vertices_i$ are the vertices adjacent to $i$ and $\alpha_{f,ij}$ is the angle in face $f$ opposite to the edge $(i,j)$ which is calculated in screen-space without foreshortening using the first fundamental form. The two constants are
\begin{align}
    m_f &:= \frac{1}{12}\sum_{\substack{i,j\in\vertices_f\\i\leq j}} (\vec{n}_f\cdot\vec{r}_i)(\vec{n}_f\cdot\vec{r}_j)  &
    b_f &:= \frac{1}{3}\sum_{i\in\vertices_f} (\vec{n}_f\cdot\vec{r}_i)
\end{align}
where $\vertices_f$ are the three vertices of a face $f$. Analogous to the pixel-based case, the linear system is underdetermined as \cref{eqn:optimality_condition} only contains depth differences, \ie adding a constant offset does not change $E_\text{Int}$.
\par
Readers familiar with the topic of mesh processing will notice the similarity to the seminal cotangent weights for the vertex Laplacian \cite{Pinkall:1993}. These weights ensure that our integration energy is independent of the triangulation. Generally, the cotan may diverge but as angles in isotropic meshes are close to $60^\circ$, the cotans are well-behaved and yield numerically stable linear systems \cite{Botsch:2010}. To our surprise and to the best of our knowledge, this discretized version of \cref{eqn:unified_integration} and its derivation for triangle meshes, has not been published yet.
\section{Implementation Details}
Our meshing and integration algorithms rely on the \textsc{SurfaceMesh} \cite{Sieger:2019} data structure and are written in \cpp. To translate between pixels and triangles we rely on the \textsc{nvdiffrast} renderer \cite{Laine:2020}. Before calculating pixel-curvatures, the normal maps are low-pass filtered with a Gaussian blur ($\sigma=\sqrt{2}$) in \textsc{OpenCV} \cite{Bradski:2000}. The sparse linear systems of the normal integration part are solved with \textsc{Eigen} \cite{Guennebaud:2010}. In all tests, we run ten iterations of mesh refinement with five iterations of tangential smoothing. The optimal edge lengths are clamped to lie within 1 and 100 pixels. A reference implementation can be found under \href{https://moritzheep.github.io/adaptive-screen-meshing/}{moritzheep.github.io/adaptive-screen-meshing}.

\begin{table}[t]
     \caption{Compression rates at three different quality settings for the five objects in the DiLiGenT-MV dataset. Compression rates are averaged over all $20$ views for each object. Results are computed from ground truth normals as well as normals obtained from photometric stereo using \cite{Li:2020} and \cite{Heep:2022}.}
    \label{tab:reduction}
    \centering
    \begin{tabular}{@{}lcrrrcrrrcrrr@{}}
        \toprule
            && \multicolumn{3}{c}{GT Normals} && \multicolumn{3}{c}{Li \cite{Li:2020}} && \multicolumn{3}{c}{Heep \cite{Heep:2022}} \\
        \cmidrule{3-5}\cmidrule{7-9}\cmidrule{11-13}
        Dataset &&  \multicolumn{1}{c}{Low}   &   \multicolumn{1}{c}{Mid}   &   \multicolumn{1}{c}{High}  &&   \multicolumn{1}{c}{Low}    &   \multicolumn{1}{c}{Mid}   &   \multicolumn{1}{c}{High}  &&   \multicolumn{1}{c}{Low}    &   \multicolumn{1}{c}{Mid}   &   \multicolumn{1}{c}{High}  \\
        \cmidrule{1-1}\cmidrule{3-5}\cmidrule{7-9}\cmidrule{11-13}
        Bear    &&  98.8\%  &  97.5\%  &    94.4\%  &&   99.0\%  &   97.7\%  &   94.7\%  &&   99.0\%  &   97.8\%  &   95.1\% \\
        Buddha  &&  96.1\%  &  92.4\%  &    82.4\%  &&   97.2\%  &   94.1\%  &   86.2\%  &&   97.0\%  &   93.7\%  &   85.4\% \\
        Cow     &&  98.9\%  &  97.6\%  &    94.3\%  &&   99.0\%  &   97.7\%  &   94.5\%  &&   97.9\%  &   95.4\%  &   89.3\% \\
        Pot2    &&  98.1\%  &  96.1\%  &    91.2\%  &&   98.4\%  &   96.6\%  &   92.1\%  &&   98.2\%  &   96.2\%  &   91.6\% \\
        Reading &&  98.2\%  &  96.0\%  &    90.7\%  &&   98.4\%  &   96.4\%  &   91.4\%  &&   95.8\%  &   91.2\%  &   79.7\% \\
    \bottomrule
    \end{tabular}
\end{table}

\begin{figure}[b]
    \includegraphics[scale=0.64]{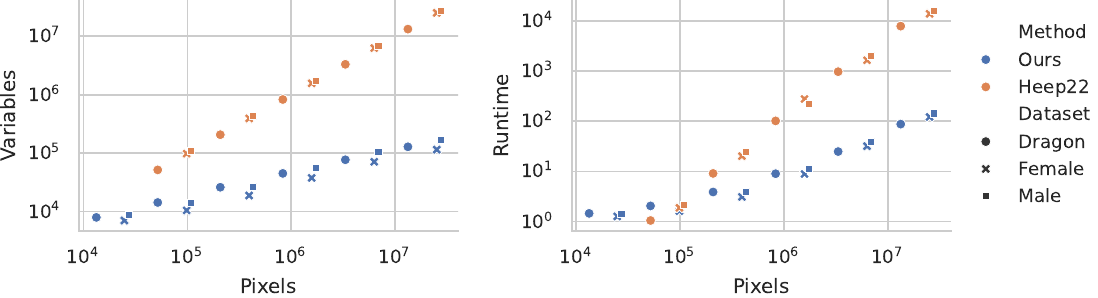}
    \caption{Runtime and compression comparison between our meshing-first and pixel-based integration for increasing image resolutions. Left: The number of free variables (pixels or vertices) as a function of foreground pixels. Right: Pixel-based integration compared to mesh first, \ie meshing and integration. Please note the log-log scale.}
    \label{fig:vertices_vs_resolution}
\end{figure}
\section{Evaluation}\label{sec:evaluation} 
Unlike previous work, our method is not bound to the image resolution. Instead, the mesh resolution is controlled by a user parameter $\epsilon$.
Therefore, we conduct all experiments at three quality settings: 'high' (\high), 'medium' (\medium) and 'low' (\low), see \cref{fig:not_a_teaser,fig:parameter_approx_error}. With the 'high' setting, we aim for accurate reconstructions, ideally matching pixel-level methods. The 'low' setting focuses on high compression rates, while the 'medium' setting balances quality and compression. 
\par
Our evaluation consists of three parts: First, we analyze the compression rates and runtime advantages of our meshing-first approach over pixel-based methods in \cref{sec:compression_runtime}. We continue by comparing the reconstruction error on samples of the DiLiGenT-MV dataset \cite{Li:2020} in \cref{sec:benchmark}. Finally, we discuss various design decisions of our algorithm in an ablation study and investigate the reliability of our single user-parameter, see \cref{sec:ablation}. The DiLiGenT-MV dataset contains $5$ objects with $20$ different views, leading to $100$ normal maps of real data for the evaluation. It is an extensive and recent dataset but lacks high-resolution normal maps. We compensate for that by testing rendered normal maps of virtual objects with resolutions of up to $8192^2$. Additional results are found in the supplemental material.
\subsection{Compression and Runtime}\label{sec:compression_runtime}
The core motivation for our meshing-first approach is to substantially reduce the number of free variables \emph{prior} to the integration step. We test all three quality settings of our screen-space meshing algorithm on all $20$ views of the DiLiGenT-MV dataset (\cref{tab:reduction}) with ground truth normals as well as estimated normals from the photometric stereo methods in \cite{Li:2020} and \cite{Heep:2022}. Despite the relatively small image size of the DiLiGenT-MV dataset, our mesh representation requires only 6-18\% of free variables for high-resolution meshes. For some low-resolution meshes, our method requires only 1\% the number of vertices compared to foreground pixels. The results are consistent across all test cases and robust even for less accurate normals. The image resolutions of the DiLiGenT-MV dataset are comparatively low (less than $1\,\text{MP}$). To obtain resolutions ranging from $512^2$ to $8192^2$ pixels, we render normal maps from three highly detailed meshes in Blender. In this application, we only use the high-resolution mesh setting.  
The resulting plots in \cref{fig:vertices_vs_resolution} suggest that the number of vertices grows sublinearly with the number of pixels, meaning that our compression factor increases at higher resolutions. For the female face, we obtain an impressive reduction by about $200$ at $8192^2$ pixels and maintain runtimes of a few minutes instead of hours. 
\begin{table}[tb]
    \centering
    \caption{Average RMSE over all $20$ views of the DiLiGenT-MV dataset for pixel-based integration \cite{Cao:2021}, {\cite{Xie:2014}}, {\cite{Heep:2022}}, a combination of pixel-based integration \cite{Cao:2021} and subsequent remeshing \cite{Dunyach:2013} as well as mesh integration of uniform and our adaptive meshes. Errors are reported in mm.} 
    \label{tab:integration_error}
    \sisetup{detect-weight=true,table-align-text-post=false}
    \newcommand{\fbest}[1]{\bfseries #1}
    \newcommand{\sbest}[1]{{\underline{\tablenum[table-format=1.2]{#1}}}}
    \begin{tabular}{c|clcS[table-format=1.2]S[table-format=1.2]S[table-format=1.2]cS[table-format=1.2]S[table-format=1.2]S[table-format=1.2]cS[table-format=1.2]S[table-format=1.2]S[table-format=1.2]cS[table-format=1.2]S[table-format=1.2]S[table-format=1.2]}
        \toprule
        \multicolumn{3}{c}{} &\multicolumn{4}{c}{Pixel-based}&& \multicolumn{3}{c}{Combined\cite{Cao:2021,Dunyach:2013}} && \multicolumn{3}{c}{Uniform} && \multicolumn{3}{c}{Ours} \\
        \cmidrule{9-11}\cmidrule{13-15}\cmidrule{17-19}
        \multicolumn{4}{c}{} & {\;\cite{Cao:2021}\;} & {\;\cite{Xie:2014}\;} & {\;\cite{Heep:2022}\;} && {\,Low\,} & {\,Mid\,} & {\,High\,} && {\,Low\,} & {\,Mid\,} & {\,High\,} && {\,Low\,} & {\,Mid\,} & {\,High\,} \\
        \cmidrule{1-3}\cmidrule{5-7}\cmidrule{9-11}\cmidrule{13-15}\cmidrule{17-19}
        \multirow{5}{*}{\rotatebox{90}{Ortho}}
        && Bear     &&  \fbest{2.97} & 3.36 & \multicolumn{1}{c}{n/a} && 3.12 & 2.99 & \sbest{2.98} && 4.39 & 4.07 & 3.74 && 3.95 & 3.65 & 3.37 \\
        && Buddha   &&  \fbest{6.74} & 6.85 & \multicolumn{1}{c}{n/a} && 6.78 & 6.76 & \sbest{6.75} && 7.79 & 7.62 & 7.41 && 7.74 & 7.54 & 7.33 \\
        && Cow      &&  \fbest{2.45} & \sbest{2.51} & \multicolumn{1}{c}{n/a} && 2.67 & 2.55 & 2.53 && 3.59 & 3.39 & 3.10 && 3.42 & 3.12 & 2.96 \\
        && Pot2     &&  \sbest{5.15} & \fbest{5.02} & \multicolumn{1}{c}{n/a} && 5.18 & 5.15 & 5.15 && 5.99 & 5.84 & 5.73 && 5.89 & 5.77 & 5.65 \\
        && Reading  &&  \sbest{6.34} & \fbest{6.05} & \multicolumn{1}{c}{n/a} && 6.37 & 6.35 & 6.34 && 7.18 & 7.00 & 6.87 && 7.08 & 6.93 & 6.83 \\
        \midrule
        \multirow{5}{*}{\rotatebox{90}{Persp}}
        && Bear    && \fbest{2.91} & \multicolumn{1}{c}{n/a} & 3.05 && 3.09 & 2.94 & \sbest{2.92} && 4.62 & 4.17 & 3.84 && 3.94 & 3.72 & 3.47 \\
        && Buddha  && \sbest{6.75} & \multicolumn{1}{c}{n/a} & \fbest{6.60} && 6.80 & 6.77 & 6.76 && 7.85 & 7.64 & 7.43 && 7.74 & 7.53 & 7.40 \\
        && Cow     && \fbest{2.35} & \multicolumn{1}{c}{n/a} & 2.51 && 2.59 & 2.46 & \sbest{2.43} && 3.71 & 3.47 & 3.21 && 3.49 & 3.24 & 3.07 \\
        && Pot2    && \fbest{4.99} & \multicolumn{1}{c}{n/a} & 5.23 && 5.04 & 4.99 & \sbest{4.99} && 6.08 & 5.97 & 5.81 && 6.04 & 5.86 & 5.76 \\
        && Reading && \fbest{6.28} & \multicolumn{1}{c}{n/a} & \sbest{6.29} && 6.33 & 6.29 & 6.29 && 7.20 & 7.03 & 6.89 && 7.19 & 6.94 & 6.85 \\
        \bottomrule
    \end{tabular}
\end{table}
\subsection{Benchmark Evaluation}\label{sec:benchmark}
We compare the reconstruction error of our adaptive mesh integration against different pixel-based approaches \cite{Xie:2014,Cao:2021,Heep:2022}. Even at the highest mesh resolution, we require on average less than 10\% of free variables. Given this level of compression, it is not surprising that pixel-based approaches perform slightly better, see \cref{tab:integration_error}. Still, discontinuities remain the main source of integration errors.
\begin{figure}[tb]
    \raggedright
    \begin{tblr}{colsep=1pt, colspec={b{.015\linewidth}b{.11\linewidth}b{.055\linewidth, rightsep=2pt}b{.11\linewidth}b{.055\linewidth, rightsep=2pt}b{.11\linewidth}b{.055\linewidth, rightsep=2pt}b{.11\linewidth}b{.055\linewidth, rightsep=4pt}b{.11\linewidth}b{.055\linewidth, rightsep=2pt}b{.02\linewidth}}, rowsep=1pt}
        &    \SetCell[c=2]{c}{\textsc{Bear}}    
        &&   \SetCell[c=2]{c}{\textsc{Buddha}}  
        &&   \SetCell[c=2]{c}{\textsc{Cow}}   
        &&   \SetCell[c=2]{c}{\textsc{Pot2}}   
        &&   \SetCell[c=2]{c}{\textsc{Reading}}    
        & \\
        \hline[white]
        \SetCell[r=4]{l}{\rotatebox{90}{\textsc{Uniform}}}
        &
        \includegraphics[width=\linewidth]{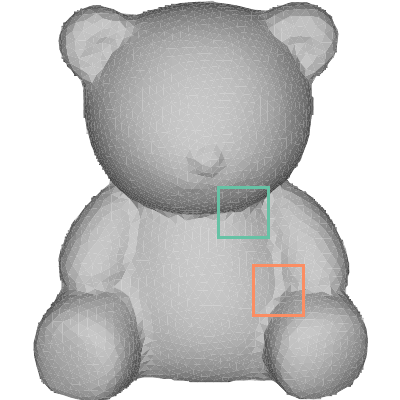}   &
        \includegraphics[width=\linewidth]{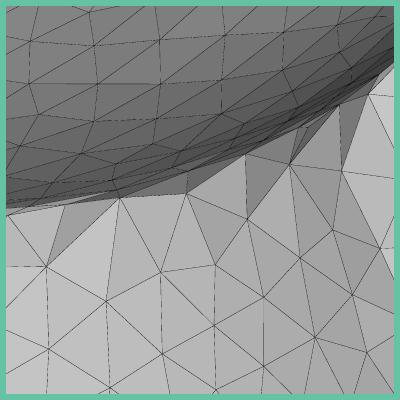} \newline
        \includegraphics[width=\linewidth]{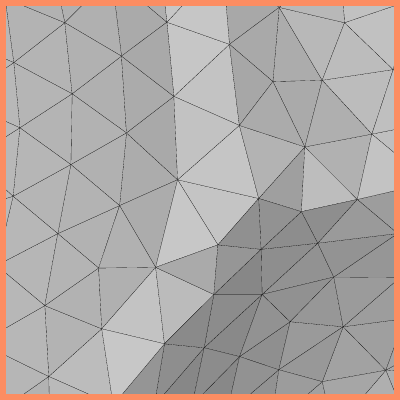} &
        \includegraphics[width=\linewidth]{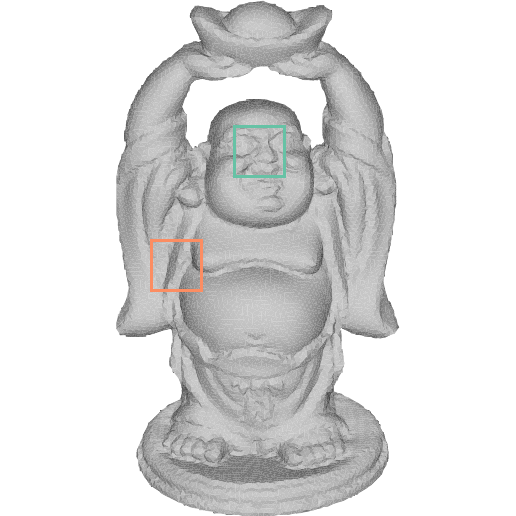}  &
        \includegraphics[width=\linewidth]{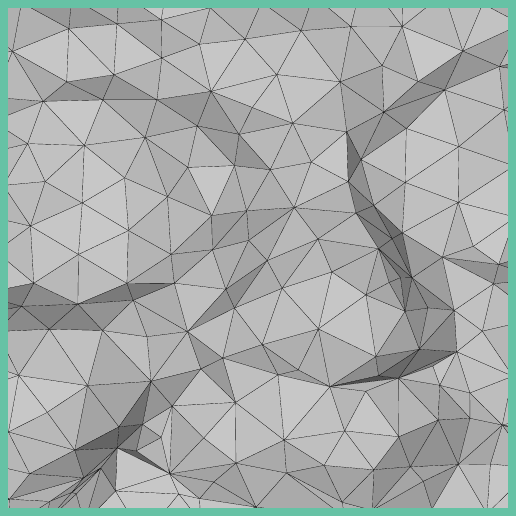} \newline
        \includegraphics[width=\linewidth]{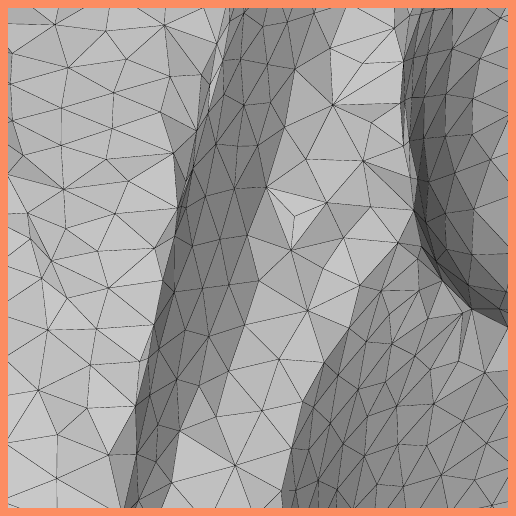} &
        \includegraphics[width=\linewidth]{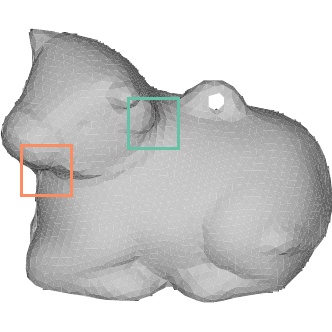} &
        \includegraphics[width=\linewidth]{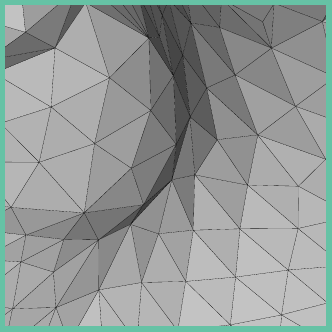} \newline
        \includegraphics[width=\linewidth]{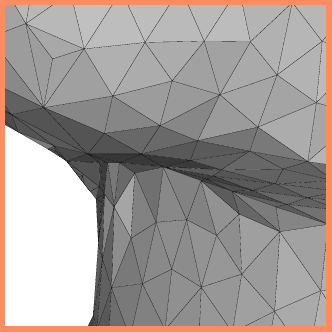} &
        \includegraphics[width=\linewidth]{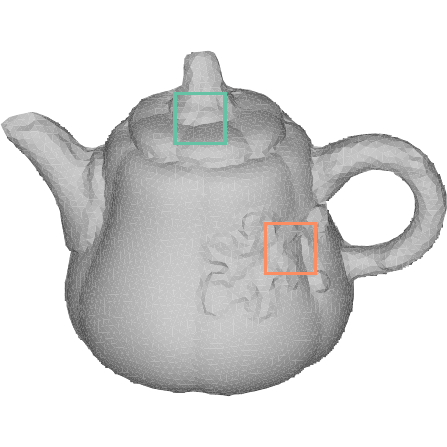}    &
        \includegraphics[width=\linewidth]{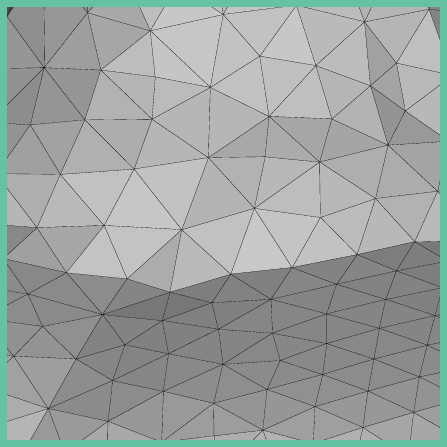} \newline
        \includegraphics[width=\linewidth]{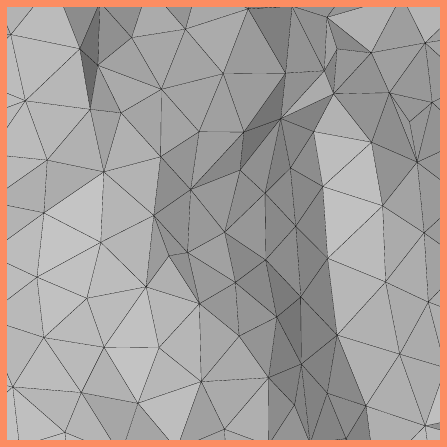} &
        \includegraphics[width=\linewidth]{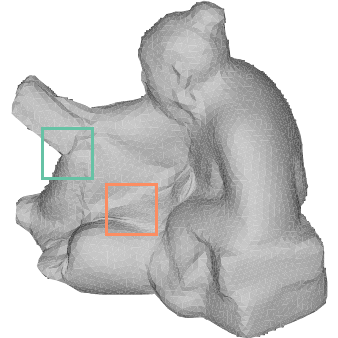} &
        \includegraphics[width=\linewidth]{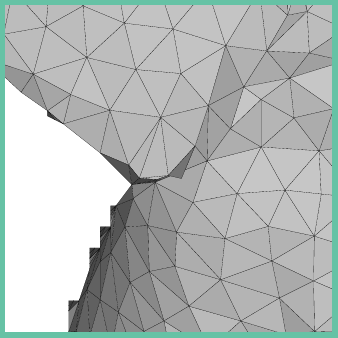} \newline
        \includegraphics[width=\linewidth]{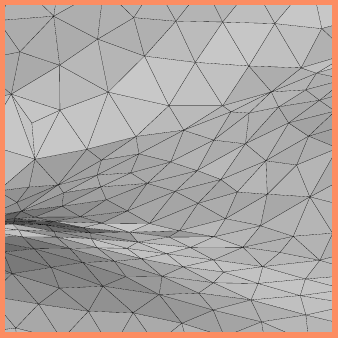} &
        \SetCell[r=8]{c,m}{{\includegraphics[scale=0.64]{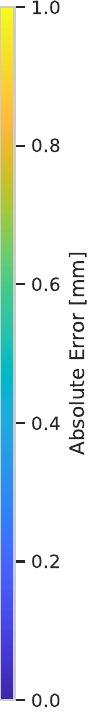}}} \\
        &    \SetCell[c=2]{c}{\footnotesize{$2613$ Vertices}}    
        &&   \SetCell[c=2]{c}{\footnotesize{$9019$ Vertices}}  
        &&   \SetCell[c=2]{c}{\footnotesize{$1823$ Vertices}}
        &&   \SetCell[c=2]{c}{\footnotesize{$3742$ Vertices}}   
        &&   \SetCell[c=2]{c}{\footnotesize{$3047$ Vertices}}
        & \\
        &
        \includegraphics[width=\linewidth]{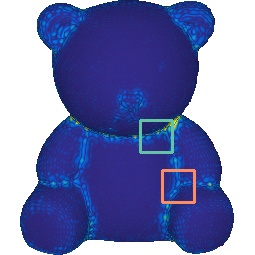}   &
        \includegraphics[width=\linewidth]{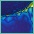} \newline
        \includegraphics[width=\linewidth]{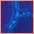} &
        \includegraphics[width=\linewidth]{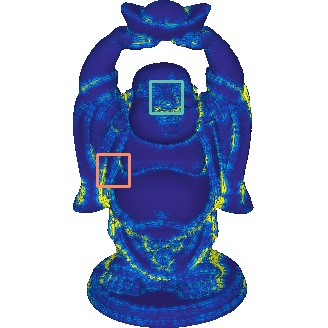}   &
        \includegraphics[width=\linewidth]{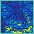} \newline
        \includegraphics[width=\linewidth]{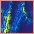} &
        \includegraphics[width=\linewidth]{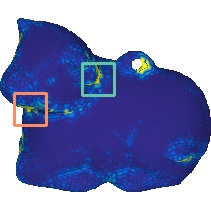}   &
        \includegraphics[width=\linewidth]{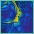} \newline
        \includegraphics[width=\linewidth]{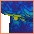} &
        \includegraphics[width=\linewidth]{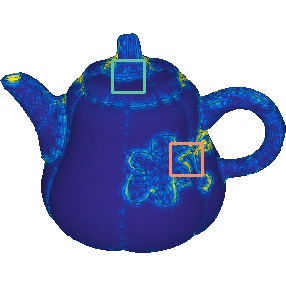}   &
        \includegraphics[width=\linewidth]{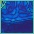} \newline
        \includegraphics[width=\linewidth]{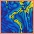} &
        \includegraphics[width=\linewidth]{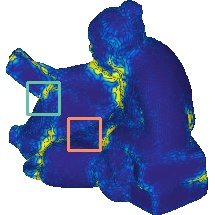}   &
        \includegraphics[width=\linewidth]{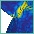} \newline
        \includegraphics[width=\linewidth]{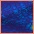} & \\
        &    \SetCell[c=2]{c}{\footnotesize{$0.112\,\text{mm}$}}    
        &&   \SetCell[c=2]{c}{\footnotesize{$0.415\,\text{mm}$}}  
        &&   \SetCell[c=2]{c}{\footnotesize{$0.282\,\text{mm}$}}
        &&   \SetCell[c=2]{c}{\footnotesize{$0.240\,\text{mm}$}}   
        &&   \SetCell[c=2]{c}{\footnotesize{$0.597\,\text{mm}$}}   
        & \\
        \cline{2-11}
        \hline[white]
        \SetCell[r=4]{l}{\rotatebox{90}{\textsc{Ours}}}
        &
        \includegraphics[width=\linewidth]{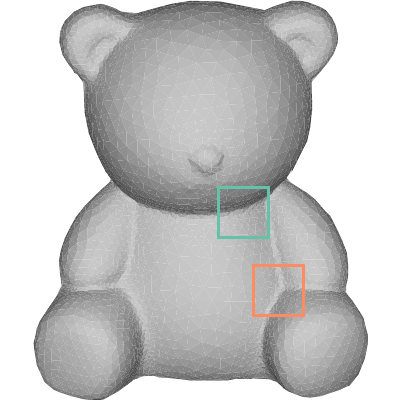}    &
        \includegraphics[width=\linewidth]{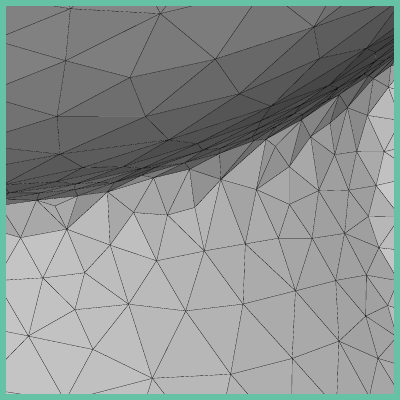} \newline
        \includegraphics[width=\linewidth]{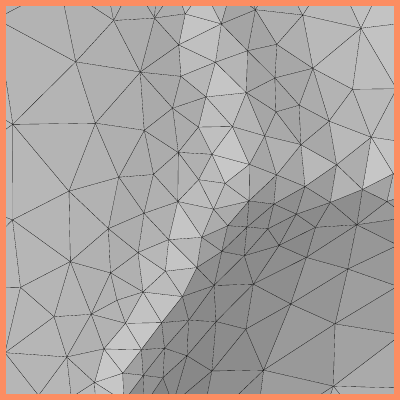} &
        \includegraphics[width=\linewidth]{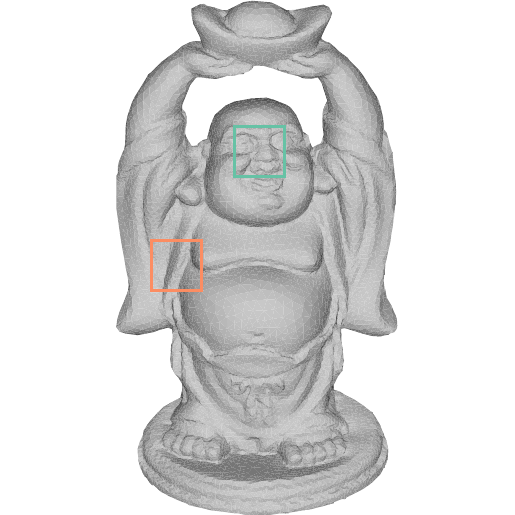}  &
        \includegraphics[width=\linewidth]{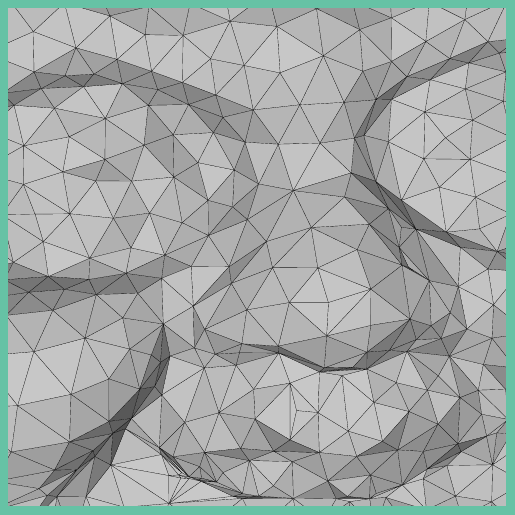} \newline
        \includegraphics[width=\linewidth]{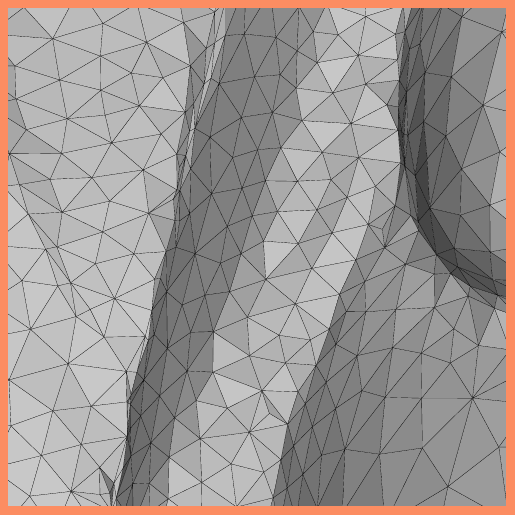} &
        \includegraphics[width=\linewidth]{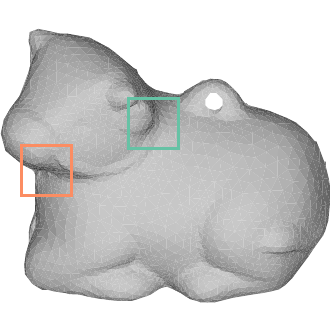} &
        \includegraphics[width=\linewidth]{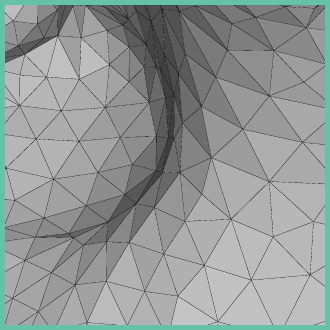} \newline
        \includegraphics[width=\linewidth]{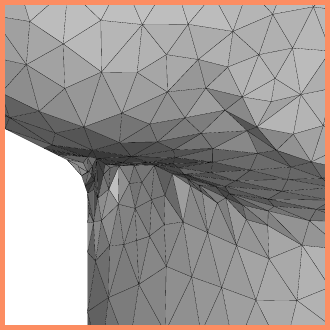} &
        \includegraphics[width=\linewidth]{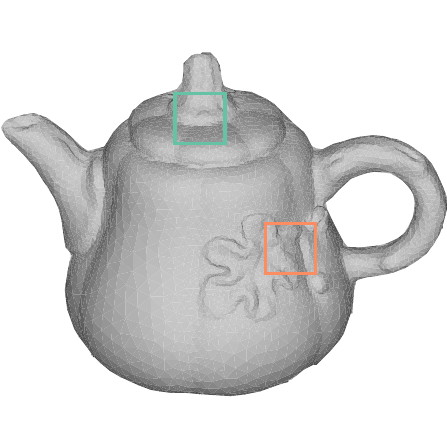}    &
        \includegraphics[width=\linewidth]{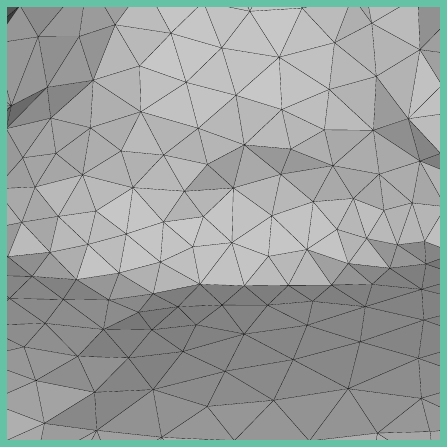} \newline
        \includegraphics[width=\linewidth]{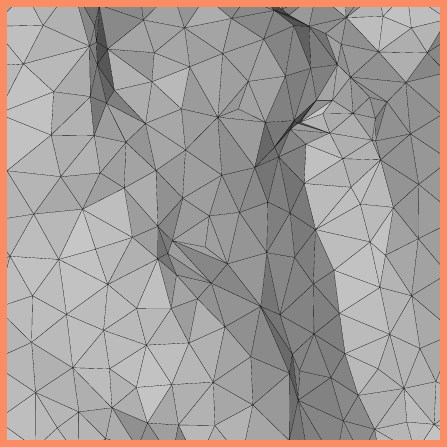} &
        \includegraphics[width=\linewidth]{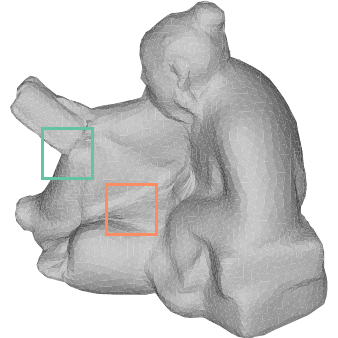} &
        \includegraphics[width=\linewidth]{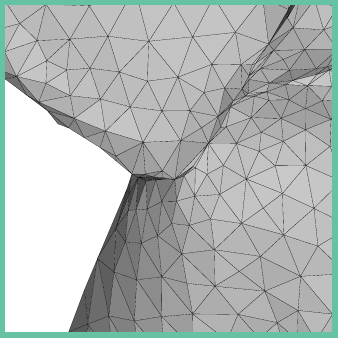} \newline
        \includegraphics[width=\linewidth]{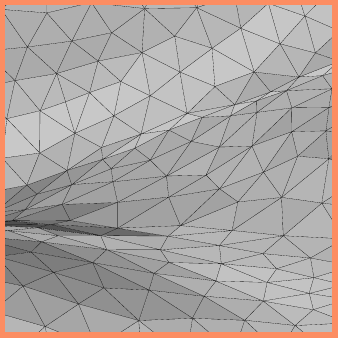} & \\
        &    \SetCell[c=2]{c}{\footnotesize{$2523$ Vertices}}    
        &&   \SetCell[c=2]{c}{\footnotesize{$8819$ Vertices}}  
        &&   \SetCell[c=2]{c}{\footnotesize{$1738$ Vertices}}
        &&   \SetCell[c=2]{c}{\footnotesize{$3590$ Vertices}}   
        &&   \SetCell[c=2]{c}{\footnotesize{$2927$ Vertices}}
        & \\
        &\includegraphics[width=\linewidth]{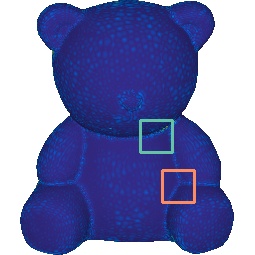}   &
        \includegraphics[width=\linewidth]{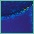} \newline
        \includegraphics[width=\linewidth]{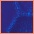} &
        \includegraphics[width=\linewidth]{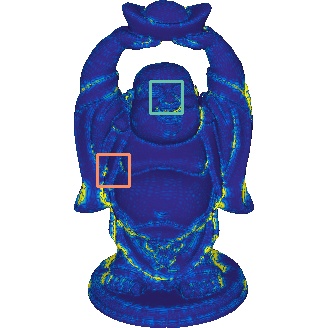}   &
        \includegraphics[width=\linewidth]{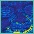} \newline
        \includegraphics[width=\linewidth]{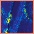} &
        \includegraphics[width=\linewidth]{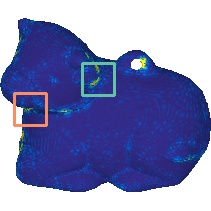}   &
        \includegraphics[width=\linewidth]{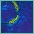} \newline
        \includegraphics[width=\linewidth]{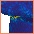} &
        \includegraphics[width=\linewidth]{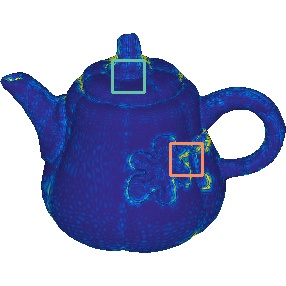}   &
        \includegraphics[width=\linewidth]{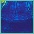} \newline
        \includegraphics[width=\linewidth]{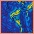} &
        \includegraphics[width=\linewidth]{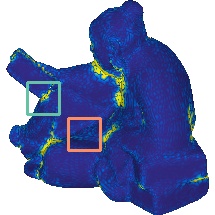}   &
        \includegraphics[width=\linewidth]{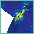} \newline
        \includegraphics[width=\linewidth]{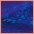} & \\
        &    \SetCell[c=2]{c}{$\mathbf{0.043\,\rm{mm}}$}    
        &&   \SetCell[c=2]{c}{$\mathbf{0.360\,\rm{mm}}$}  
        &&   \SetCell[c=2]{c}{$\mathbf{0.173\,\rm{mm}}$}
        &&   \SetCell[c=2]{c}{$\mathbf{0.187\,\rm{mm}}$}   
        &&   \SetCell[c=2]{c}{$\mathbf{0.476\,\rm{mm}}$}   
    \end{tblr}
    \caption{Comparison between uniform (\emph{top}) and adaptive screen-space meshing (\emph{bottom}). Despite up to $5\%$ more vertices, the RMSE is always higher for the uniform meshes. Wireframes are rendered as vector graphics.}
    \label{fig:adapted_vs_uniform}
\end{figure}

\begin{figure}[tb]
    \centering
    \begin{subfigure}{.17\linewidth}
        \includegraphics[width=\linewidth, trim={0 20 0 20}, clip]{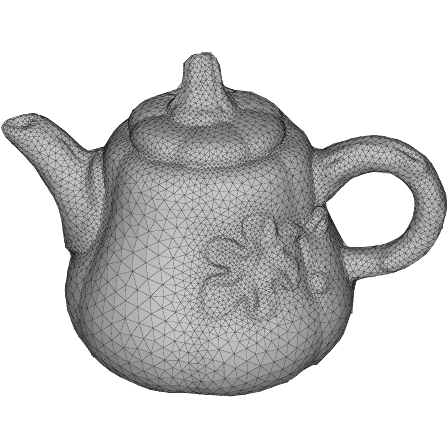}
        \includegraphics[width=\linewidth, trim={0 20 0 20}, clip]{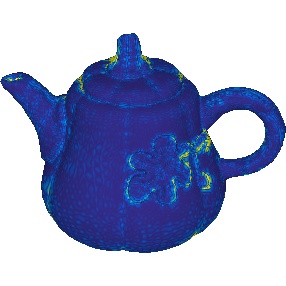}
        \caption*{High}
    \end{subfigure}
    \begin{subfigure}{.17\linewidth}
        \includegraphics[width=\linewidth, trim={0 20 0 20}, clip]{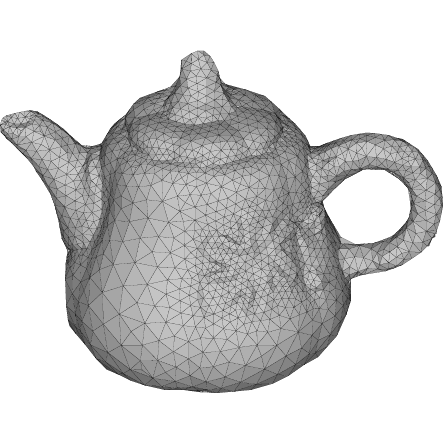}
        \includegraphics[width=\linewidth, trim={0 20 0 20}, clip]{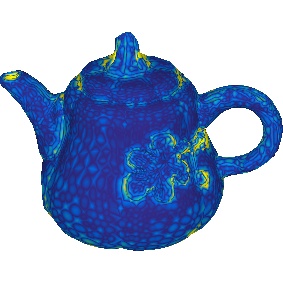}
        \caption*{Medium}
    \end{subfigure}
    \begin{subfigure}{.17\linewidth}
        \raggedleft
        \includegraphics[width=\linewidth, trim={0 20 0 20}, clip]{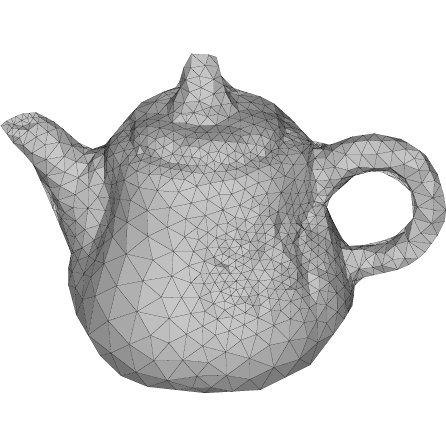}
        \includegraphics[width=\linewidth, trim={0 20 0 20}, clip]{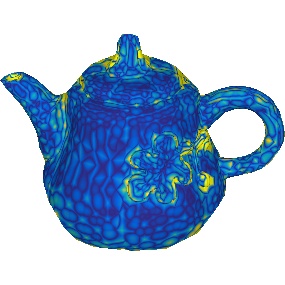}
        \caption*{Low}
    \end{subfigure}
    \begin{subfigure}{.46\linewidth}
        \includegraphics[scale=0.64]{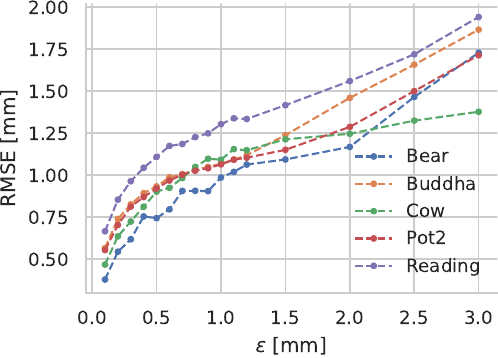}
    \end{subfigure}
    \caption{Impact analysis of the user-parameter. Left: Obtained high (\high), medium (\medium) and low (\low) resolution meshes (as vector graphics) with respective error maps. Right: Average RMSE over all $20$ views as a function of the user-parameter $\epsilon$.}
    \label{fig:parameter_approx_error}
\end{figure}
\subsection{Ablation Study}\label{sec:ablation}
Our mesh refinement algorithm consists of three iterative steps, all described and evaluated in previous work or textbooks \cite{Botsch:2004,Botsch:2010}. The remaining design choice of the algorithm is our locally adaptive over a uniform triangulation, see \cref{fig:adapted_vs_uniform}. We create uniform meshes as a baseline by directly setting a constant optimal edge length such that the resulting mesh roughly matches the vertex count of its adaptive counterpart. To exclude possible side-effects of the surface integration, the 2D meshes are non-rigidly fitted to ground truth depth maps. Thus, the remaining RMSE quantifies the error introduced by the mesh representation alone. Although our locally adaptive triangulations always contain equal or fewer vertices than the uniform triangulations, they still achieve a lower alignment error in all cases, see \cref{fig:adapted_vs_uniform}.
\par
Lastly, our method relies on a single user parameter to control the density of the mesh, namely $\epsilon$, the permitted error of the surface. Throughout our evaluations, we consistently reported the results for three settings the parameter. In \cref{fig:parameter_approx_error}, we set $\epsilon\in[0.1\,\text{mm},3\,\text{mm}]$ and non-rigidly fit the 2D meshes to ground truth depth maps. The results show a strong correlation between the intended and achieved approximation error. After a dataset-dependent offset, the achieved error grows approximately linearly with $\epsilon$. Additional results in the supplementary material suggest that this linear relation is still present after integration.
\section{Conclusion \& Limitations}\label{sec:conclusion}
We proposed a novel normal integration method, where normal maps are converted into triangle meshes \emph{before} the integration. A single user parameter controls the mesh resolution. We showed that our data structure is substantially sparser, but well-suited to obtain detailed reconstructions. Our results show that the mesh representation is especially advantageous for high-resolution normal maps, where it facilitates enormous speed-ups. As with all work on normal integration, the resulting surfaces are only unique up to a global scale and absolute depth remains ill-posed. Furthermore, we do not handle depth discontinuities implicitly at the moment. For triangle meshes, we would need to control the mesh topology and align the triangle edges with the discontinuities along the surface. Developing such a method for our mesh-based integration is not straightforward and is an interesting research direction. Alternatively, depth discontinuities could be precomputed and introduced as small holes into the foreground masks. Anisotropic meshes to improve the handling of sharp corners are another interesting direction for future work. Here, skinny triangles might be a drawback as they may cause numerical instabilities due to diverging cotangents. Finally, our method can be easily integrated into most existing photometric stereo pipelines.
\section*{Acknowledgements}
The "David Head" by \href{https://skfb.ly/oKvzW}{1d\_inc} and the "Football Medal2 - PhotoCatch" by \href{https://skfb.ly/oGXrA}{Moshe Caine} were licensed under CC BY 4.0.
This work has been funded by the Deutsche Forschungsgemeinschaft (DFG, German Research Foundation) under Germany's Excellence Strategy, EXC-2070 -- 390732324 (PhenoRob).
\FloatBarrier
\bibliographystyle{splncs04}
\bibliography{eccv24}

\begin{thebibliography}{10}
\providecommand{\url}[1]{\texttt{#1}}
\providecommand{\urlprefix}{URL }
\providecommand{\doi}[1]{https://doi.org/#1}

\bibitem{Ackermann:2015}
Ackermann, J., Goesele, M.: A {{Survey}} of {{Photometric Stereo Techniques}}.
  Foundations and Trends in Computer Graphics and Vision  \textbf{9}(3-4),
  149--254 (2015). \doi{10.1561/0600000065}

\bibitem{Alliez:2005}
Alliez, P., de~Verdi{\`e}re, {\'E}.C., Devillers, O., Isenburg, M.: Centroidal
  {{Voronoi}} diagrams for isotropic surface remeshing. Graphical Models
  \textbf{67}(3),  204--231 (May 2005). \doi{10.1016/j.gmod.2004.06.007}

\bibitem{Botsch:2004}
Botsch, M., Kobbelt, L.: A remeshing approach to multiresolution modeling. In:
  Proceedings of the 2004 {{Eurographics}}/{{ACM SIGGRAPH}} Symposium on
  {{Geometry}} Processing. pp. 185--192 (2004)

\bibitem{Botsch:2010}
Botsch, M., Kobbelt, L., Pauly, M., Alliez, P., L{\'e}vy, B.: Polygon Mesh
  Processing. CRC press (2010)

\bibitem{Bradski:2000}
Bradski, G.: The {{OpenCV}} library. Dr. Dobb's Journal of Software Tools
  (2000)

\bibitem{Cao:2022}
Cao, X., Santo, H., Shi, B., Okura, F., Matsushita, Y.: Bilateral {{Normal
  Integration}}. In: Computer {{Vision}} -- {{ECCV}} 2022, vol. 13661, pp.
  552--567. Springer Nature Switzerland, Cham (2022).
  \doi{10.1007/978-3-031-19769-7_32}

\bibitem{Cao:2021}
Cao, X., Shi, B., Fumio, O., Matsushita, Y.: Normal {{Integration}} via
  {{Inverse Plane Fitting With Minimum Point-to-Plane Distance}}. In:
  Proceedings of the {{IEEE}}/{{CVF Conference}} on {{Computer Vision}} and
  {{Pattern Recognition}} ({{CVPR}}). pp. 2382--2391 (Jun 2021)

\bibitem{Chen:2004}
Chen, L., Xu, J.c.: Optimal delaunay triangulations. Journal of Computational
  Mathematics pp. 299--308 (2004)

\bibitem{Chen:2012}
Chen, Z., Cao, J., Wang, W.: Isotropic {{Surface Remeshing Using Constrained
  Centroidal Delaunay Mesh}}. Computer Graphics Forum  \textbf{31}(7),
  2077--2085 (Sep 2012). \doi{10.1111/j.1467-8659.2012.03200.x}

\bibitem{Du:1999}
Du, Q., Faber, V., Gunzburger, M.: Centroidal {{Voronoi Tessellations}}:
  {{Applications}} and {{Algorithms}}. SIAM Review  \textbf{41}(4),  637--676
  (Jan 1999). \doi{10.1137/S0036144599352836}

\bibitem{Du:2007}
Du, Z., {Robles-Kelly}, A., Lu, F.: Robust {{Surface Reconstruction}} from
  {{Gradient Field Using}} the {{L1 Norm}}. In: 9th {{Biennial Conference}} of
  the {{Australian Pattern Recognition Society}} on {{Digital Image Computing
  Techniques}} and {{Applications}} ({{DICTA}} 2007). pp. 203--209 (Dec 2007).
  \doi{10.1109/DICTA.2007.4426797}

\bibitem{Dunyach:2013}
Dunyach, M., Vanderhaeghe, D., Barthe, L., Botsch, M.: Adaptive remeshing for
  real-time mesh deformation. In: Eurographics 2013 - {{Short Papers}}. The
  Eurographics Association (2013)

\bibitem{Durou:2007}
Durou, J.D., Courteille, F.: Integration of a {{Normal Field}} without
  {{Boundary Condition}}. In: Proceedings of the {{First International
  Workshop}} on {{Photometric Analysis For Computer Vision-PACV}} 2007.
  pp.~8--p. INRIA (2007)

\bibitem{Dziuba:2023}
Dziuba, M., Jarsky, I., Efimova, V., Filchenkov, A.: Image {{Vectorization}}: A
  {{Review}} (Jun 2023). \doi{10.48550/arXiv.2306.06441}

\bibitem{Gan:2018}
Gan, J., Wilbert, A., Thorm{\"a}hlen, T., Drescher, P., Hagens, R.: Multi-view
  photometric stereo using surface deformation. The Visual Computer
  \textbf{34}(11),  1551--1561 (Nov 2018). \doi{10.1007/s00371-017-1430-5}

\bibitem{Garland:1997}
Garland, M., Heckbert, P.S.: Surface simplification using quadric error
  metrics. In: Proceedings of the 24th Annual Conference on {{Computer}}
  Graphics and Interactive Techniques - {{SIGGRAPH}} '97. pp. 209--216. ACM
  Press, Not Known (1997). \doi{10.1145/258734.258849}

\bibitem{Gotardo:2015}
Gotardo, P.F.U., Simon, T., Sheikh, Y., Matthews, I.: Photogeometric {{Scene
  Flow}} for {{High-Detail Dynamic 3D Reconstruction}}. In: Proceedings of the
  {{IEEE International Conference}} on {{Computer Vision}} ({{ICCV}}). pp.
  846--854. IEEE, Santiago, Chile (Dec 2015). \doi{10.1109/ICCV.2015.103}

\bibitem{Griwodz:2021}
Griwodz, C., Gasparini, S., Calvet, L., Gurdjos, P., Castan, F., Maujean, B.,
  Lillo, G.D., Lanthony, Y.: {{AliceVision Meshroom}}: {{An}} open-source
  {{3D}} reconstruction pipeline. In: Proceedings of the 12th {{ACM Multimedia
  Systems Conference}} - {{MMSys}} '21. ACM Press (2021).
  \doi{10.1145/3458305.3478443}

\bibitem{Guennebaud:2010}
Guennebaud, G., Jacob, B., et~al.: Eigen v3 (2010)

\bibitem{Heep:2022}
Heep, M., Zell, E.: {{ShadowPatch}}: {{Shadow Based Segmentation}} for
  {{Reliable Depth Discontinuities}} in {{Photometric Stereo}}. Computer
  Graphics Forum  \textbf{41}(7),  635--646 (Oct 2022). \doi{10.1111/cgf.14707}

\bibitem{Horn:1986}
Horn, B.K., Brooks, M.J.: The variational approach to shape from shading.
  Computer Vision, Graphics, and Image Processing  \textbf{33}(2),  174--208
  (1986)

\bibitem{Khan:2020}
Khan, D., Plopski, A., Fujimoto, Y., Kanbara, M., Jabeen, G., Zhang, Y.J.,
  Zhang, X., Kato, H.: Surface remeshing: {{A}} systematic literature review of
  methods and research directions. IEEE Transactions on Visualization and
  Computer Graphics  \textbf{28}(3),  1680--1713 (2020)

\bibitem{Laine:2020}
Laine, S., Hellsten, J., Karras, T., Seol, Y., Lehtinen, J., Aila, T.: Modular
  primitives for high-performance differentiable rendering. ACM Transactions on
  Graphics (TOG)  \textbf{39}(6),  1--14 (2020)

\bibitem{Li:2020}
Li, M., Zhou, Z., Wu, Z., Shi, B., Diao, C., Tan, P.: Multi-view photometric
  stereo: {{A}} robust solution and benchmark dataset for spatially varying
  isotropic materials. IEEE Transactions on Image Processing  \textbf{29},
  4159--4173 (2020)

\bibitem{Liao:2012}
Liao, Z., Hoppe, H., Forsyth, D., Yu, Y.: A subdivision-based representation
  for vector image editing. IEEE Transactions on Visualization and Computer
  Graphics  \textbf{18}(11),  1858--1867 (2012)

\bibitem{Ma:2022}
Ma, X., Zhou, Y., Xu, X., Sun, B., Filev, V., Orlov, N., Fu, Y., Shi, H.:
  Towards layer-wise image vectorization. In: Proceedings of the {{IEEE}}/{{CVF
  Conference}} on {{Computer Vision}} and {{Pattern Recognition}} ({{CVPR}}).
  pp. 16314--16323 (2022)

\bibitem{Pinkall:1993}
Pinkall, U., Polthier, K.: Computing discrete minimal surfaces and their
  conjugates. Experimental mathematics  \textbf{2}(1),  15--36 (1993)

\bibitem{Queau:2018}
Qu{\'e}au, Y., Durou, J.D., Aujol, J.F.: Normal {{Integration}}: {{A Survey}}.
  Journal of Mathematical Imaging and Vision  \textbf{60}(4),  576--593 (2018)

\bibitem{Schonberger:2016a}
Sch{\"o}nberger, J.L., Frahm, J.M.: Structure-from-{{Motion Revisited}}. In:
  Conference on {{Computer Vision}} and {{Pattern Recognition}} ({{CVPR}})
  (2016)

\bibitem{Schonberger:2016}
Sch{\"o}nberger, J.L., Zheng, E., Pollefeys, M., Frahm, J.M.: Pixelwise {{View
  Selection}} for {{Unstructured Multi-View Stereo}}. In: European
  {{Conference}} on {{Computer Vision}} ({{ECCV}}) (2016)

\bibitem{Sieger:2019}
Sieger, D., Botsch, M.: The polygon mesh processing library (2019)

\bibitem{Xie:2015}
Xie, W., Dai, C., Wang, C.C.: Photometric stereo with near point lighting:
  {{A}} solution by mesh deformation. In: Proceedings of the {{IEEE
  Conference}} on {{Computer Vision}} and {{Pattern Recognition}} ({{CVPR}}).
  pp. 4585--4593 (2015)

\bibitem{Xie:2019}
Xie, W., Wang, M., Wei, M., Jiang, J., Qin, J.: Surface {{Reconstruction}} from
  {{Normals}}: {{A Robust DGP-based Discontinuity Preservation Approach}}. In:
  Proceedings of the {{IEEE}}/{{CVF Conference}} on {{Computer Vision}} and
  {{Pattern Recognition}} ({{CVPR}}). pp. 5328--5336 (2019)

\bibitem{Xie:2014}
Xie, W., Zhang, Y., Wang, C.C., Chung, R.C.K.: Surface-from-gradients: {{An}}
  approach based on discrete geometry processing. In: Proceedings of the {{IEEE
  Conference}} on {{Computer Vision}} and {{Pattern Recognition}} ({{CVPR}}).
  pp. 2195--2202 (2014)

\bibitem{Yi:2018}
Yi, R., Liu, Y.J., He, Y.: Delaunay mesh simplification with differential
  evolution. ACM Transactions on Graphics (TOG)  \textbf{37}(6),  1--12 (2018)

\bibitem{Zhao:2023}
Zhao, T., Bus{\'e}, L., {Cohen-Steiner}, D., Boubekeur, T., Thiery, J.M.,
  Alliez, P.: Variational shape reconstruction via quadric error metrics. In:
  {{ACM SIGGRAPH}} 2023 {{Conference Proceedings}}. pp. 1--10 (2023)

\bibitem{Zhou:2020}
Zhou, M., Ding, Y., Ji, Y., Young, S.S., Yu, J., Ye, J.: Shape and
  {{Reflectance Reconstruction}} using {{Concentric Multi-Spectral Light
  Field}}. IEEE Transactions on Pattern Analysis and Machine Intelligence
  \textbf{42}(7),  1594--1605 (2020)

\bibitem{Zhu:2020}
Zhu, D., Smith, W.A.P.: Least {{Squares Surface Reconstruction}} on {{Arbitrary
  Domains}}. In: Computer {{Vision}} -- {{ECCV}} 2020, vol. 12367, pp.
  530--545. Springer International Publishing, Cham (2020).
  \doi{10.1007/978-3-030-58542-6_32}

\end{thebibliography}
\end{document}